\documentclass[11pt]{article}
\usepackage[preprint]{acl}
\usepackage{times,latexsym}
\usepackage[T1]{fontenc}
\usepackage[utf8]{inputenc}
\ifdefined\XeTeXversion
\catcode"2019=13 \def’{'}
\catcode"2014=13 \def—{---}
\catcode"2013=13 \def–{--}
\else
\DeclareUnicodeCharacter{2019}{'}
\DeclareUnicodeCharacter{2014}{---}
\DeclareUnicodeCharacter{2013}{--}
\fi
\usepackage{cuted}
\usepackage{multicol}
\usepackage{capt-of} 
\usepackage{dblfloatfix}
\usepackage{microtype,inconsolata}
\usepackage{graphicx,booktabs,amsmath,amssymb,multirow,array}
\usepackage{enumitem,float}

\newcommand{\obs}{OBS}

\newcommand{\acc}{\operatorname{Acc}}
\title{How Reliable Are System-One Decisions in Medicine? A Comparative Study of Jev, Jev-like Models, and LLMs}
\author{
  Yanrui Du\textsuperscript{1,2}\thanks{Equal contribution.},
  Danyang Zhao\textsuperscript{1}\footnotemark[1],
  Sendong Zhao\textsuperscript{1},
  Wenhao Li\textsuperscript{1} \\
  \textbf{
    Yibo Gao\textsuperscript{1},
    Haochun Wang\textsuperscript{1},
    Zhuoer Xu\textsuperscript{2},
    Shiwen Cui\textsuperscript{2},
    Bing Qin\textsuperscript{1}
  } \\
  \textsuperscript{1}Harbin Institute of Technology \\
  \textsuperscript{2}Ant Group
}
\begin{document}
\maketitle
\begin{abstract}

Jev has attracted considerable attention for its rapid, generalizable decision-making, yet its medical reliability remains underexplored. 
To address this gap, our study compares various types of system-one decision models across 15 medical datasets comprising 8,742 instances under a unified protocol. 
Beyond accuracy, we assess decision stability and the ability to recognize unreliable decisions. 
Our results show that Jev rivals leading models on primarily text-driven tasks but is less effective on numerically driven tasks, revealing a distinct capability boundary.
Under question-format variations, Jev maintains strong decision consistency and small accuracy fluctuations. StartLux-27B exhibits comparable consistency, while D3-27B achieves higher overall accuracy and a lower mean decision reversal rate.
But these strengths do not extend to reliable abstention awareness: most evaluated models show limitations in both recognizing missing correct answers and using optional abstention.
Moreover, we find that Jev’s probability-based answer selection can be improved by changing how decisions are scored: normalizing option-wise binary scores yields higher retained accuracy than choice scoring.
Taken together, our findings provide early guidance for improving and applying decision models in medicine.
\end{abstract}


\section{Introduction}

Medical decision-making demands reliability beyond accuracy, as errors and inconsistencies under uncertainty can compromise patient care. 
Reliable models should not only make accurate decisions, but also remain consistent under meaning-preserving format changes, recognize missing correct answers, and abstain when appropriate. 
Their probability scores should further help distinguish reliable decisions from errors~\citep{hager,metamed}. 
These requirements make medicine a demanding setting for evaluating both the quality of model decisions and the extent to which they can be trusted.

\begin{figure*}[!t]
\centering
\includegraphics[width=0.90\textwidth]{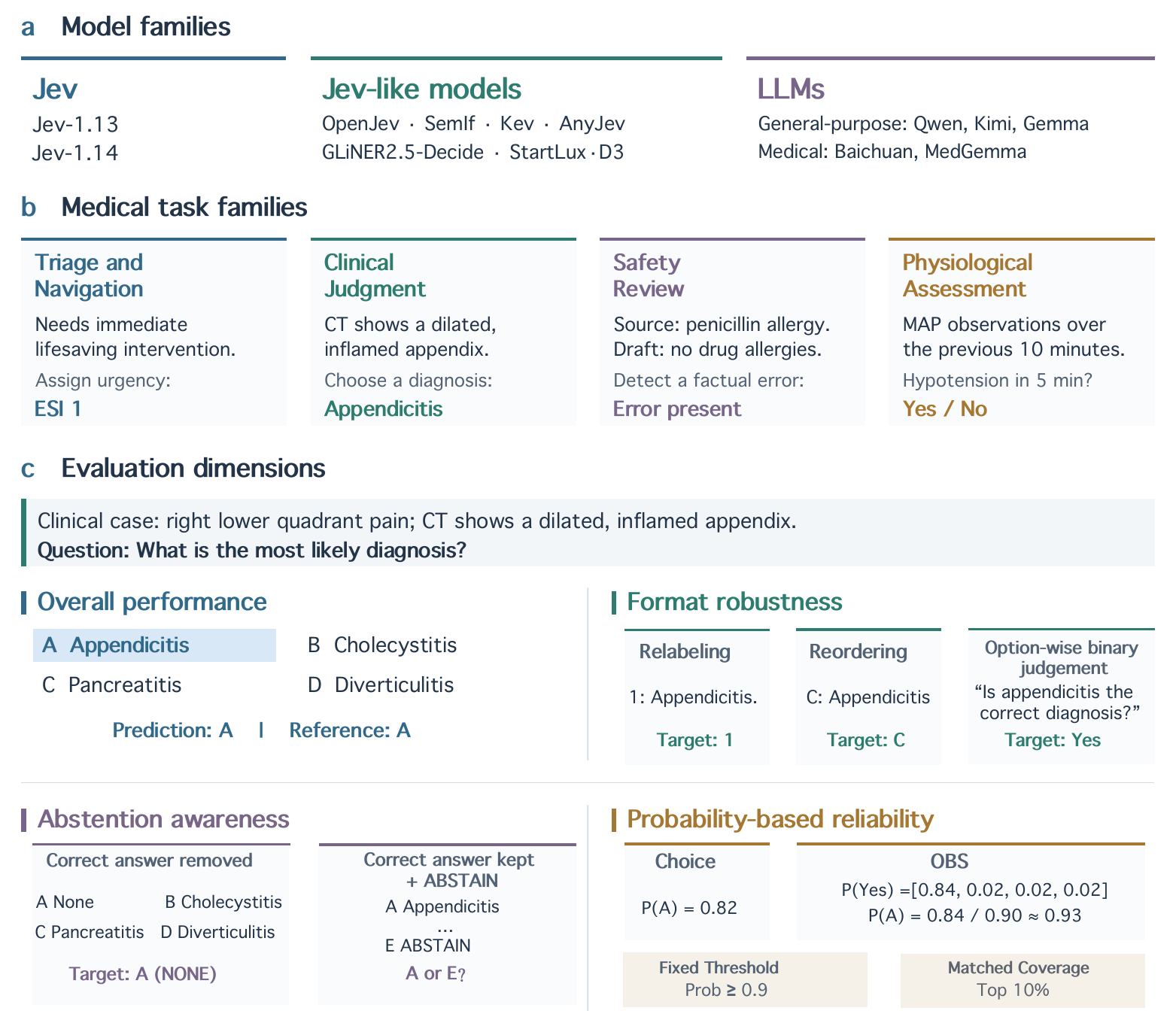}
\caption{\textbf{Overview of the comparative evaluation.} Three model families, four medical task families, and four evaluation dimensions. A shared clinical judgment case illustrates performance, format changes, missing-answer recognition, optional abstention, and probability-based selection.}
\label{fig:framework}
\end{figure*}

Traditional decision models typically pair pretrained backbones with task-specific classification heads, requiring additional training or adaptation when tasks change~\citep{devlin-etal-2019-bert}. 
Jev offers a different approach: it directly scores candidate decisions specified at inference time without task-specific training, enabling rapid decision-making across heterogeneous tasks~\citep{typesafe,jev_general}. 
Together with emerging Jev-like models, it represents a growing class of \emph{system-one decision models} that make direct decisions without elicited reasoning traces. 
Yet broader task generality does not guarantee reliable decisions, particularly in demanding medical settings. 
Existing medical evaluations of Jev have primarily focused on accuracy within a limited range of tasks~\citep{jev_medicine,jev_radiology}, leaving its broader decision reliability insufficiently explored. 
This gap motivates a systematic comparison across medical tasks and model families that goes beyond accuracy to examine decision consistency, abstention awareness, and probability-based reliability.

To this end, we systematically compare two Jev service versions, ten Jev-like model configurations, and eleven general-purpose and medical large language models (LLMs) across 15 medical datasets comprising 8,742 instances. 
All models are evaluated under a system-one protocol that requires direct decisions without elicited reasoning traces. 
As illustrated in Figure~\ref{fig:framework}, the evaluation spans four medical task families, covering both primarily text-driven and numerically driven decisions. 
We measure overall performance using accuracy and macro-F1, alongside three complementary dimensions of reliability. 
\textbf{Format robustness} examines decision consistency under option relabeling, option reordering, and conversion to option-wise binary judgments. 
\textbf{Abstention awareness} assesses whether models recognize missing correct answers and abstain effectively when given the option. 
\textbf{Probability-based reliability} examines whether higher decision probabilities identify more accurate decisions at fixed probability thresholds and matched coverage.

Our results reveal distinct strengths and limitations across system-one medical decision models. Jev rivals leading LLMs on primarily text-driven tasks but remains substantially weaker on numerically driven Physiological Assessment. It also maintains strong format robustness, with Jev-1.14 achieving the lowest accuracy-change variance. D3-27B attains the highest overall accuracy and macro-F1 and the lowest mean decision reversal rate, demonstrating that open-source alternatives can match or exceed Jev on these dimensions. However, these strengths do not ensure reliable abstention awareness, which remains limited even among the most accurate and consistent models. For probability-based selection, normalized option-wise binary scoring (OBS) improves Jev's retained accuracy at every tested threshold and matched coverage level in both versions, but does not provide the same benefit for all models. These findings highlight task suitability, abstention behavior, and scoring formulation as distinct considerations for reliable medical decision-making.
Overall, our contributions can be summarized as:
\begin{itemize}[leftmargin=*,noitemsep,topsep=2pt]
\item \textbf{Comprehensive evaluation.} We establish a unified evaluation of system-one medical decision-making, comparing 23 models across 15 datasets, 8,742 instances, and 4 medical task families.
\item \textbf{Reliability characterization.} We evaluate reliability beyond aggregate accuracy through format robustness and abstention awareness, revealing clear task-dependent capability boundaries and persistent limitations in abstention behavior.
\item \textbf{Probability-based selection.} We systematically compare direct Choice scoring with normalizing option-wise scores (OBS), identifying OBS as a more reliable strategy for selecting Jev’s answers.
\end{itemize}

\section{Related Work}
\subsection{Generalizable Decision Models}
Jev scores instruction-conditioned candidates across tasks rather than generating unrestricted text~\citep{typesafe,jev_general}. Jev-like alternatives explore related interfaces through different architectures and readouts, while prior studies examine scientific decisions, probability semantics, and presentation sensitivity~\citep{jev_science,sys1cal,jev_names,jev_adv}. Medical evaluations cover examination questions and radiology factuality but remain limited in task coverage and cross-model comparisons~\citep{korean_pilot,jev_medicine,jev_radiology}. Our evaluation jointly examines task performance and reliability across model families; Appendix~\ref{app:related_extended} provides additional context and model references.

\subsection{Medical Evaluation and Reliability}
Medical benchmarks extend from knowledge examinations to records, medical actions, and health conversations~\citep{medqa,medxpert,medhelm,hager,healthbench}. Reliability studies additionally examine input and option sensitivity~\citep{medfuzz,order,selectionbias}, answer availability and abstention~\citep{know,metamed}, and selective decision-making and calibration~\citep{selective,guo}. These complementary perspectives motivate our evaluation beyond aggregate accuracy. Appendix~\ref{app:related_extended} details their relationship to the present candidate-based framework.

\begin{table*}[t]
\centering\small\setlength{\tabcolsep}{7pt}
\begin{tabular*}{\textwidth}{@{\extracolsep{\fill}}lrrrrr@{}}
\toprule
Model & \shortstack{A.Triage and\\Navigation} & \shortstack{B.Clinical\\Judgment} & \shortstack{C.Safety\\Review} & \shortstack{D.Physiological\\Assessment} & Overall \\
\midrule
\multicolumn{6}{@{}l}{\textit{Jev}} \\
Jev-1.13 & 55.9 / 50.8 & 74.9 / 74.3 & 59.8 / 58.9 & 35.9 / 27.2 & 61.6 / 58.7 \\
Jev-1.14 & 54.4 / 50.1 & 76.8 / 76.2 & 58.9 / 58.0 & 39.4 / 32.0 & 62.3 / 59.7 \\
\midrule
\multicolumn{6}{@{}l}{\textit{General-purpose LLMs}} \\
Qwen3.5-0.8B & 26.3 / 11.7 & 39.7 / 35.0 & 38.3 / 27.9 & 53.4 / 33.8 & 37.6 / 27.2 \\
Qwen3.5-4B & 43.2 / 35.7 & 60.4 / 59.7 & 51.4 / 45.9 & 54.8 / 39.1 & 53.2 / 47.8 \\
Qwen3.5-9B & 49.5 / 46.5 & 64.4 / 62.7 & 48.6 / 46.2 & 51.4 / 40.2 & 55.5 / 52.1 \\
Qwen3.6-27B & 57.3 / \textbf{54.4} & 72.9 / 71.5 & 59.9 / 56.3 & 56.7 / 50.9 & 64.0 / 61.2 \\
Qwen3.8-27B & \textbf{57.5} / 50.5 & 72.9 / 72.5 & 58.8 / 58.8 & 56.8 / \textbf{52.8} & 63.8 / 61.3 \\
Gemma4-12B-IT & 49.6 / 44.9 & 69.9 / 69.0 & 56.5 / 52.4 & 40.8 / 32.5 & 57.9 / 54.4 \\
Gemma4-31B-IT & 54.5 / 46.8 & \textbf{80.1} / \textbf{79.8} & 60.0 / 55.0 & \textbf{58.1} / 50.4 & 66.3 / 62.1 \\
Kimi-K2.6 & 38.0 / 32.7 & 65.7 / 61.0 & 54.6 / 51.4 & 51.3 / 49.2 & 54.2 / 50.0 \\
\midrule
\multicolumn{6}{@{}l}{\textit{Medical LLMs}} \\
MedGemma-4B-IT & 40.3 / 30.9 & 56.3 / 52.7 & 42.9 / 41.1 & 51.4 / 29.1 & 48.7 / 41.4 \\
MedGemma-27B-IT & 40.8 / 35.1 & 63.6 / 61.7 & 52.4 / 53.0 & 51.7 / 29.3 & 53.7 / 48.5 \\
Baichuan-M3-235B & 48.0 / 42.8 & 65.9 / 64.1 & 50.6 / 46.5 & 52.5 / 49.9 & 56.3 / 53.0 \\
\midrule
\multicolumn{6}{@{}l}{\textit{Jev-like models}} \\
GLiNER2.5-Decide & 22.9 / 9.4 & 28.7 / 17.5 & 39.8 / 21.6 & 51.8 / 28.3 & 32.5 / 17.6 \\
SemIf & 44.4 / 37.2 & 58.8 / 57.4 & 50.0 / 45.8 & 46.3 / 42.2 & 51.5 / 47.7 \\
Kev-4B & 47.6 / 42.8 & 63.8 / 63.2 & 55.9 / 52.7 & 44.4 / 39.0 & 55.3 / 52.4 \\
AnyJev & 38.3 / 28.2 & 53.8 / 51.9 & 43.4 / 41.9 & 55.1 / 46.3 & 47.8 / 42.8 \\
OpenJev & 54.8 / 48.3 & 71.1 / 70.0 & 59.3 / 55.5 & 48.1 / 39.9 & 61.3 / 57.3 \\
StartLux-0.8B & 36.9 / 24.9 & 49.1 / 47.3 & 47.1 / 43.0 & 52.3 / 33.4 & 45.9 / 38.6 \\
StartLux-4B & 49.7 / 44.4 & 69.2 / 68.9 & 60.0 / 58.3 & 42.5 / 32.4 & 58.6 / 55.4 \\
StartLux-9B & 51.6 / 45.4 & 73.1 / 72.4 & 57.0 / 56.8 & 41.9 / 35.5 & 60.0 / 57.2 \\
StartLux-27B & 54.4 / 47.9 & 78.9 / 78.1 & 62.0 / 62.4 & 47.6 / 34.7 & 64.8 / 61.1 \\
D3-27B & 55.8 / 48.9 & 79.2 / 78.9 & \textbf{63.0} / \textbf{63.1} & 55.3 / 38.2 & \textbf{66.5} / \textbf{62.3} \\
\bottomrule
\end{tabular*}

\caption{Original-choice \textbf{accuracy / macro-F1} (\%), averaged equally across datasets. Bold indicates column-wise best scores. More detailed dataset-level results appear in Figure~\ref{fig:accuracy} in Appendix~\ref{app:performance}.}
\vspace{-0.5cm}
\label{tab:overall}
\end{table*}

\section{Preliminary}
\label{sec:preliminary}
\label{sec:benchmarks}
We evaluate 8,742 entries from 15 datasets in four task families. Each input contains evidence, instructions, and candidate decisions.

\paragraph{A: Triage and Navigation.}
Patient messages and triage records support care-seeking and urgency decisions (3,095 entries).

\paragraph{B: Clinical Judgment.}
Medical narratives support diagnosis, management, coding, and rule- or evidence-based judgments (2,501 entries).

\paragraph{C: Safety Review.}
Medical text and patient constraints support error detection, document-risk assessment, and medication selection (1,812 entries).

\paragraph{D: Physiological Assessment.}
Numerical features and time-stamped observations support state classification and short-horizon event assessment (1,334 entries).

Dataset sources, references, and input--output examples appear in Appendices~\ref{app:preliminary_details}--\ref{app:examples}.

\paragraph{Models and decision protocol.}
\label{sec:models_protocols}
\label{sec:systemone}
Both Jev versions, ten Jev-like configurations, and eleven general-purpose or medical LLMs decide without elicited reasoning or tools. They receive matched task content, excluding reference answers. Choice selects the highest-scoring candidate. LLMs use the forced assistant prefill ``\texttt{The answer is \textbackslash boxed\{}'' and candidate-normalized next-token probabilities. Option-wise binary scoring (OBS) separately judges each candidate and normalizes affirmative probabilities across candidates. Model sources, readouts, and label conventions are detailed in Appendix~\ref{app:models}.

\section{Overall Performance Evaluation}
\label{sec:performance}
The overall performance helps establish the baseline for subsequent reliability analyses.

\subsection{Evaluation Setup}
We evaluate the original Choice using accuracy and macro-F1 (\%). Accuracy accepts any reference answer, and macro-F1 uses unique-reference entries and each dataset's fixed label inventory. Task-family and overall scores average dataset-level scores equally, over the constituent datasets and all 15 datasets, respectively. Reference handling is detailed in Appendix~\ref{app:data}.

\subsection{Results and Analysis}

\paragraph{Competitive text-driven decisions, weaker numerical assessment.}
Jev approaches leading LLMs across the three primarily text-driven families in Table~\ref{tab:overall}. In Clinical Judgment, its accuracy of $74.9$--$76.8\%$ and macro-F1 of $74.3$--$76.2\%$ exceed Qwen3.8-27B's $72.9/72.5$ but trail Gemma4-31B-IT's $80.1/79.8$. In Safety Review, Jev achieves accuracy close to Gemma4-31B-IT ($58.9$--$59.8\%$ versus $60.0\%$) with higher macro-F1 ($58.0$--$58.9\%$ versus $55.0\%$). This competitiveness does not extend to numerically driven Physiological Assessment, where Jev achieves accuracy of $35.9$--$39.4\%$ and macro-F1 of $27.2$--$32.0\%$, compared with Gemma4-31B-IT's $58.1/50.4$. Across the two versions, CTG-21 accuracy remains at $20.7$--$27.6\%$, below the cohort's $53.5\%$ majority-class baseline, while balanced VitalDB accuracy is $51.2\%$ for both. The task boundary is therefore more informative than Jev's overall accuracy of $61.6$--$62.3\%$ and macro-F1 of $58.7$--$59.7\%$.

\paragraph{Jev-like alternatives are competitive, but task profiles differ.}

D3-27B achieves the highest overall accuracy and macro-F1 ($66.54/62.34$), slightly exceeding Gemma4-31B-IT ($66.34/62.13$), and leads Safety Review ($63.02/63.09$). StartLux-27B also exceeds both Jev versions overall ($64.83/61.11$), whereas OpenJev remains close to Jev-1.13 ($61.34/57.31$). These aggregate rankings do not imply uniform superiority. D3-27B reaches $55.30/38.21$ in Physiological Assessment, above Jev but below Gemma4-31B-IT's $58.10/50.37$. The dataset-level results in Figure~\ref{fig:accuracy} further qualify these averages. Jev exceeds $94\%$ accuracy on ICD-proxy and MIMIC-CDM but struggles on MedXpertQA and SCT; within Triage and Navigation, broader care-seeking decisions are easier than five-level ESI classification. Model selection therefore requires attention to task-specific strengths rather than overall rankings alone.

\section{Format Robustness Evaluation}
\label{sec:robustness}

Beyond accuracy, we examine format robustness in terms of how consistently models respond
when the same medical evidence and candidate meanings are presented
through different request formats. 
We assess consistency at two levels: aggregate performance and individual decisions.

\subsection{Evaluation Setup}

To examine these levels, we use three paired interventions.
Relabeling tests sensitivity to candidate labels, and reordering tests sensitivity to candidate order. Both preserve the
original decision formulation. Option-wise binary scoring (OBS) tests
sensitivity to reformulating a joint choice as separate candidate
judgments, using the protocol described in
Section~\ref{sec:systemone}. Together, these interventions examine
presentation invariance and consistency across decision formulations.
Table~\ref{tab:transformations} illustrates the interventions using
the same synthetic case as Figure~\ref{fig:framework}.

\begin{table*}[!t]
\centering
\small
\setlength{\tabcolsep}{5pt}
\begin{tabular}{
>{\raggedright\arraybackslash}p{2.05cm}
>{\raggedright\arraybackslash}p{6.75cm}
>{\raggedright\arraybackslash}p{3.2cm}
>{\raggedright\arraybackslash}p{2.55cm}}
\toprule
Format & Candidate presentation & Probability readout & Identity alignment \\
\midrule
Original
& A: appendicitis; B: cholecystitis;
  C: pancreatitis; D: diverticulitis.
& \multirow{3}{3.2cm}{\raggedright
  Joint candidate probabilities $p_k$;
  select $\arg\max_k p_k$.}
& -- \\
Relabeling
& 1: appendicitis; 2: cholecystitis;
  3: pancreatitis; 4: diverticulitis.
&
& 1$\to$A, 2$\to$B, 3$\to$C, 4$\to$D. \\
Reordering
& A: pancreatitis; B: diverticulitis;
  C: appendicitis; D: cholecystitis.
&
& A$\to$C, B$\to$D, C$\to$A, D$\to$B. \\
Binary (OBS)
& Original A--D candidates retained. For each candidate:
  ``Is [candidate] the correct diagnosis?'' Yes / No.
& Affirmative $q_k$; normalize
  $\tilde p_k=q_k/\sum_jq_j$;
  select $\arg\max_k\tilde p_k$.
& -- \\
\bottomrule
\end{tabular}
\caption{Paired format interventions for one synthetic case:
CT shows an inflamed appendix; what is the most likely diagnosis?
Every row retains the same evidence and candidate meanings.
The final column maps presented candidate identities to original identities.}
\label{tab:transformations}
\end{table*}

\paragraph{Performance stability.}
We measure changes in accuracy and its variation across formats.
For dataset $d$ and transformation $t$,
\begin{align}
\Delta_{dt}
&=100(\acc_{dt}-\acc_{d0}), \\
\overline{\Delta}_d
&=\frac14\sum_{t=0}^{3}\Delta_{dt}, \\
V_{\Delta}
&=\frac1D\sum_{d=1}^{D}\frac14
  \sum_{t=0}^{3}
  (\Delta_{dt}-\overline{\Delta}_d)^2.
\label{eq:format_variance}
\end{align}
Here, $t=0$ denotes the original format, $\Delta_{d0}=0$,
and $D=15$. We report dataset-mean accuracy changes in percentage
points and the mean within-dataset variance $V_{\Delta}$ in squared
percentage points.

\paragraph{Decision consistency.}
We measure whether individual answers change after semantic alignment.
With alignment mapping $\pi_t$ and $N_d$ entries in dataset $d$,
\begin{align}
F_{dt}
&=\frac1{N_d}\sum_{i=1}^{N_d}
  \mathbf{1}
  [\hat y_{i0}\ne\pi_t(\hat y_{it})], \\
\overline F
&=\frac1{3D}\sum_{d=1}^{D}\sum_{t=1}^{3}F_{dt}.
\label{eq:format_reversal}
\end{align}
We report reversal rates for each transformation and their mean as
percentages. This captures answer changes that may be masked by stable
aggregate accuracy.

\begin{table*}[t]
\centering
\small
\setlength{\tabcolsep}{4pt}
\begin{tabular*}{\textwidth}{@{\extracolsep{\fill}}lrrrrrrrr@{}}
\toprule

Model & $\Delta$ Rel. & $\Delta$ Reord. & $\Delta$ OBS
& $V_{\Delta}$ & $F_{\mathrm{Rel}}$ & $F_{\mathrm{Reo}}$
& $F_{\mathrm{OBS}}$ & $\overline F$ \\

\midrule
\multicolumn{9}{@{}l}{\textit{Jev}} \\
Jev-1.13 & +0.45 & -0.03 & +0.05 & 1.70 & 3.9 & 6.3 & 8.3 & 6.2 \\
Jev-1.14 & +0.19 & +0.03 & +0.19 & 1.09 & 3.4 & 6.5 & 7.1 & 5.7 \\
\midrule
\multicolumn{9}{@{}l}{\textit{General-purpose LLMs}} \\
Qwen3.6-27B & -0.17 & -0.34 & -3.55 & 18.29 & 5.4 & 9.2 & 16.1 & 10.2 \\
Qwen3.8-27B & +0.40 & -0.20 & -3.84 & 19.79 & 5.1 & 10.3 & 16.9 & 10.7 \\
Gemma4-31B-IT & +0.42 & -0.59 & -2.84 & 22.23 & 3.5 & 6.3 & 12.4 & 7.4 \\
\midrule
\multicolumn{9}{@{}l}{\textit{Medical LLMs}} \\
Baichuan-M3-235B & +0.74 & +1.38 & -6.16 & 21.04 & 6.4 & 16.6 & 33.9 & 19.0 \\
\midrule
\multicolumn{9}{@{}l}{\textit{Jev-like models}} \\
OpenJev & -0.68 & +0.66 & -2.73 & 6.85 & 8.2 & 13.6 & 19.3 & 13.7 \\
StartLux-4B & -0.27 & -0.75 & -0.39 & 1.26 & 4.8 & 8.6 & 9.5 & 7.6 \\
StartLux-9B & -0.23 & +0.38 & -0.28 & 1.31 & 4.4 & 8.1 & 9.2 & 7.2 \\
StartLux-27B & -0.08 & -0.14 & -0.25 & 6.45 & 2.9 & 6.5 & 10.0 & 6.5 \\
D3-27B & -0.24 & +0.33 & -0.41 & 2.62 & 2.5 & 4.5 & 7.3 & 4.7 \\
\bottomrule
\end{tabular*}

\caption{Format robustness across all 15 datasets, with equal
dataset weights. $\Delta$ Rel., $\Delta$ Reord., and $\Delta$ OBS
denote accuracy changes (percentage points) under relabeling, reordering, and OBS, respectively.
$V_{\Delta}$ denotes mean within-dataset accuracy variance
(squared percentage points).
$F_{\mathrm{Rel}}$, $F_{\mathrm{Reo}}$, and $F_{\mathrm{OBS}}$
denote prediction reversal rates (\%) under the corresponding
interventions and $\overline F$ is their mean. Alongside both Jev versions, the three general LLMs, one medical LLM, and five Jev-like models with highest overall accuracy in Table~\ref{tab:overall} are shown. Table~\ref{tab:robustness_full} reports all 23 models.}
\label{tab:robustness}

\end{table*}

\subsection{Results and Analysis}

\paragraph{Jev combines stable accuracy with consistent decisions.}

Across relabeling, reordering, and OBS, Jev's accuracy changes remain within $0.45$ percentage points of the original scores (Table~\ref{tab:robustness}). Its mean within-dataset variance is $1.09$--$1.70$~pp$^2$, below every evaluated LLM, with Jev-1.14 achieving the lowest variance among all configurations. Jev also maintains low mean reversal rates of $5.69$--$6.17\%$, compared with $7.41\%$ for Gemma4-31B-IT and $10.75\%$ for Qwen3.8-27B. Thus, its stable aggregate accuracy is accompanied by strong individual decision consistency, although D3-27B achieves a lower mean reversal rate.

\paragraph{Reformulation exposes differences hidden by baseline accuracy.}
OBS leaves Jev's accuracy nearly unchanged but reduces Gemma4-31B-IT, Qwen3.6-27B, and Qwen3.8-27B accuracy by $2.84$, $3.55$, and $3.84$ points, respectively. Kimi-K2.6 instead gains $5.42$ points (Table~\ref{tab:robustness_full}). The effect of changing the decision formulation is therefore model-dependent, even among models with competitive original-choice scores.

\paragraph{D3 and StartLux narrow the robustness gap.}

D3-27B achieves the lowest mean reversal rate ($4.74\%$), followed by Jev-1.14 ($5.69\%$), Jev-1.13 ($6.17\%$), and StartLux-27B ($6.47\%$). Its advantage is not uniform across robustness measures. D3-27B has a higher accuracy-change variance than Jev-1.14 ($2.62$ versus $1.09$~pp$^2$), and a slightly higher OBS reversal rate ($7.27\%$ versus $7.11\%$). Other alternatives remain less consistent, with OpenJev reaching $13.71\%$ mean reversal and SemIf, AnyJev, and GLiNER2.5-Decide exceeding $32\%$ (Table~\ref{tab:robustness_full}). These results distinguish aggregate performance stability from decision consistency and show substantial variation among Jev-like models.
\section{Abstention Awareness Evaluation}
\label{sec:boundaries}

We next examine whether models recognize when no correct option is
available and whether optional abstention helps them avoid incorrect
decisions when a correct option remains present.

\subsection{Evaluation Setup}

We assess abstention awareness through two direct-choice interventions.
\textbf{Correct-answer replacement} substitutes ``none of the above''
(NONE) for the correct candidate while preserving its position and
all other options. Selecting NONE is correct in the modified task.
\textbf{Optional abstention} retains all original candidates and adds
ABSTAIN to examine whether withholding decisions improve retained
accuracy at the cost of coverage.
In the example in Figure~\ref{fig:framework}, the former replaces
appendicitis in option A with NONE, whereas the latter retains all
four diagnoses and adds ABSTAIN as option E.

Both experiments focus on Clinical Judgment, where clearly
distinguishable alternatives support interpretable evaluation.
Correct-answer replacement covers four datasets (2,382 entries)
with unique accepted answers and more than three options, whereas
optional abstention covers all six datasets (2,501 entries).

\begin{table*}[t]
\centering
\small
\setlength{\tabcolsep}{5pt}
\begin{tabular*}{\textwidth}{@{\extracolsep{\fill}}lrrrrrr@{}}
\toprule
Model & Original Acc & NONE Acc & $\Delta$ Acc & Slot flip & Correctness flip & \shortstack[r]{NONE Acc\\(orig. correct)} \\
\midrule
\multicolumn{7}{@{}l}{\textit{Jev}} \\
Jev-1.13 & 77.6 & 65.6 & -12.0 & 18.4 & 16.2 & 70.1 \\
Jev-1.14 & 77.7 & 66.3 & -11.4 & 18.2 & 16.0 & 71.0 \\
\midrule
\multicolumn{7}{@{}l}{\textit{General-purpose LLMs}} \\
Qwen3.6-27B & 78.9 & 69.0 & -9.9 & 16.2 & 15.4 & 73.7 \\
Qwen3.8-27B & 78.6 & 68.4 & -10.2 & 17.5 & 15.8 & 72.8 \\
Gemma4-31B-IT & 82.3 & 65.9 & -16.4 & 22.6 & 22.0 & 70.5 \\
\midrule
\multicolumn{7}{@{}l}{\textit{Medical LLMs}} \\
Baichuan-M3-235B & 72.1 & 51.1 & -21.0 & 28.1 & 25.8 & 54.8 \\
\midrule
\multicolumn{7}{@{}l}{\textit{Jev-like models}} \\
OpenJev & 75.8 & 64.7 & -11.1 & 20.6 & 18.2 & 70.2 \\
StartLux-4B & 71.3 & 60.0 & -11.2 & 20.1 & 16.9 & 64.2 \\
StartLux-9B & 75.7 & 60.9 & -14.8 & 20.8 & 19.6 & 64.5 \\
StartLux-27B & 83.3 & 66.8 & -16.5 & 22.4 & 21.2 & 69.5 \\
D3-27B & 82.5 & 67.0 & -15.5 & 20.7 & 19.3 & 70.1 \\
\bottomrule
\end{tabular*}

\caption{Correct-answer replacement across four Clinical Judgment
datasets (2,382 paired entries).
Accuracy and reversal rates are percentages. $\Delta$ Acc is
NONE accuracy minus original accuracy in percentage points.
NONE Acc (orig. correct) measures accuracy after replacement
on items each model originally answered correctly. Model selection follows Table~\ref{tab:robustness}; complete results appear in Table~\ref{tab:none_full}.}
\label{tab:none}
\end{table*}

\begin{table*}[t]
\centering
\small
\setlength{\tabcolsep}{4pt}
\begin{tabular*}{\textwidth}{@{\extracolsep{\fill}}lrrrrrr@{}}
\toprule

& \multicolumn{2}{c}{Original Choice}
& \multicolumn{2}{c}{With Abstention}
& \multicolumn{2}{c}{Gains} \\
\cmidrule(lr){2-3}
\cmidrule(lr){4-5}
\cmidrule(l){6-7}
Model
& $A^0_{\mathrm{all}}$
& $A^0_S$
& $R$
& $A^+_S$
& $\Delta$
& $\Delta_S$ \\
\midrule
\multicolumn{7}{@{}l}{\textit{Jev}} \\
Jev-1.13 & 74.9 & 75.5 & 4.5 & 75.2 & +0.3 & -0.3 \\
Jev-1.14 & 76.8 & 77.4 & 4.4 & 77.3 & +0.5 & -0.1 \\
\midrule
\multicolumn{7}{@{}l}{\textit{General-purpose LLMs}} \\
Qwen3.6-27B & 72.9 & 73.5 & 2.8 & 73.2 & +0.3 & -0.3 \\
Qwen3.8-27B & 72.9 & 73.4 & 2.3 & 74.1 & +1.3 & +0.8 \\
Gemma4-31B-IT & 80.1 & 80.9 & 4.2 & 80.7 & +0.6 & -0.1 \\
\midrule
\multicolumn{7}{@{}l}{\textit{Medical LLMs}} \\
Baichuan-M3-235B & 65.9 & 66.3 & 1.7 & 67.6 & +1.8 & +1.3 \\
\midrule
\multicolumn{7}{@{}l}{\textit{Jev-like models}} \\
OpenJev & 71.1 & 71.9 & 4.1 & 72.7 & +1.6 & +0.8 \\
StartLux-4B & 69.2 & 70.2 & 6.0 & 69.8 & +0.7 & -0.3 \\
StartLux-9B & 73.1 & 73.8 & 4.7 & 73.4 & +0.2 & -0.4 \\
StartLux-27B & 78.9 & 79.7 & 5.6 & 78.2 & -0.7 & -1.5 \\
D3-27B & 79.2 & 79.7 & 2.9 & 78.9 & -0.4 & -0.8 \\
\bottomrule
\end{tabular*}

\caption{Optional abstention across six Clinical Judgment datasets
(2,501 entries).
$A_{\mathrm{all}}^0$ and $A_S^0$ are original accuracy on the full
cohort and retained subset. $R$ is the mean abstention rate and
$A_S^+$ the mean retained accuracy (all in \%).
Gains $\Delta$ and $\Delta^S$ are in percentage points, computed
before rounding. Model selection follows Table~\ref{tab:robustness}; complete results appear in Table~\ref{tab:abstention_B_all}.}
\vspace{-0.3cm}
\label{tab:abstention_B}
\end{table*}

\paragraph{Evaluation metrics.}
For correct-answer replacement, we report NONE accuracy and its
percentage-point change from original accuracy on matched items.
We additionally measure \emph{conditional NONE accuracy}, reported
as NONE Acc (orig. correct), by restricting each dataset to items
the model originally answered correctly.
This measures whether models recognize missing correct answers
on questions they initially answered successfully.
Moreover, Slot flip measures changes in the selected option position,
whereas correctness flip counts both correct-to-incorrect
and incorrect-to-correct transitions.
The latter captures the total change in correctness, unlike
accuracy differences, which reflect the net balance of
improvements and errors.

For optional abstention, let $S_d$ denote the retained subset of
dataset $d$, $N_d$ its original size, and $\mathcal D_S$ the datasets
with nonempty retained subsets.
We report the mean abstention rate and retained accuracy
\begin{align}
R &= \frac{1}{D_B}\sum_{d=1}^{D_B}
\left(1-\frac{|S_d|}{N_d}\right),\\
A_S^+ &= \frac{1}{|\mathcal D_S|}
\sum_{d\in\mathcal D_S}\frac{1}{|S_d|}
\sum_{i\in S_d}\mathbf{1}[\hat y_{di}^+\in G_{di}],
\label{eq:abstention_metrics}
\end{align}
where $D_B=6$, $\hat y_{di}^+$ is the decision after adding ABSTAIN,
and $G_{di}$ is the accepted reference set.
Mean coverage is $1-R$.
Accuracy changes are measured against original decisions on the
full cohort and the same retained subsets
\begin{equation}
\Delta=100(A_S^+-A_{\mathrm{all}}^0),
\qquad
\Delta^S=100(A_S^+-A_S^0).
\label{eq:abstention_gains}
\end{equation}
The former captures the gain associated with abstention, while
the latter controls for sample selection.
All metrics use equal dataset weights, excluding empty subsets
from retained-accuracy means.
Exact retained counts and dataset-mean macro-F1 are reported in
Table~\ref{tab:abstention} in the appendix.

\subsection{Results and Analysis}

\paragraph{Task competence does not ensure missing-answer recognition.}

Correct-answer replacement reduces accuracy for 21 of 23 configurations (Tables~\ref{tab:none} and~\ref{tab:none_full}). Jev loses $11.4$--$12.0$ percentage points, while D3-27B falls from $82.49\%$ to $66.97\%$, a decline of $15.52$ points despite its leading overall performance and decision consistency. Gemma4-31B-IT and StartLux-27B likewise lose $16.4$ and $16.5$ points. Failures persist on originally correct items, where conditional NONE accuracy is only $70.1$--$71.0\%$ for Jev and $70.07\%$ for D3-27B. Kev-4B and GLiNER2.5-Decide are the only models whose overall replacement accuracy improves. Missing-answer recognition therefore remains a distinct limitation rather than a direct consequence of task competence.

\paragraph{Optional abstention offers limited gains at a coverage cost.}

Jev abstains on $4.4$--$4.5\%$ of entries on average across datasets, yet retained accuracy improves by only $0.3$--$0.5$ percentage points over the original full-cohort baseline (Tables~\ref{tab:abstention_B} and~\ref{tab:abstention_B_all}). Gemma4-31B-IT similarly gains $0.61$ points with $4.21\%$ abstention. D3-27B and StartLux-27B instead lose $0.37$ and $0.66$ points while abstaining on $2.95\%$ and $5.58\%$, respectively. Kev-4B gains $4.06$ points but withholds $23.87\%$ of decisions. An explicit abstention option therefore does not consistently improve the accuracy of retained decisions.

\paragraph{Selection benefits can mask worse retained answers.}

Comparisons on identical retained subsets separate selection effects from changes to the decisions themselves. Jev's post-abstention accuracy is $0.1$--$0.3$ percentage points below its original accuracy on these subsets. D3-27B, StartLux-27B, and Kev-4B likewise lose $0.81$, $1.49$, and $1.05$ points against their same-subset controls. Effective abstention must both identify unreliable decisions and preserve the quality of retained ones. Neither strong task performance nor format robustness guarantees these two abilities.
\raggedbottom
\section{Probability-Based Reliability}
\label{sec:confidence}
\begin{figure*}[!t]
\centering
\includegraphics[width=0.88\textwidth]{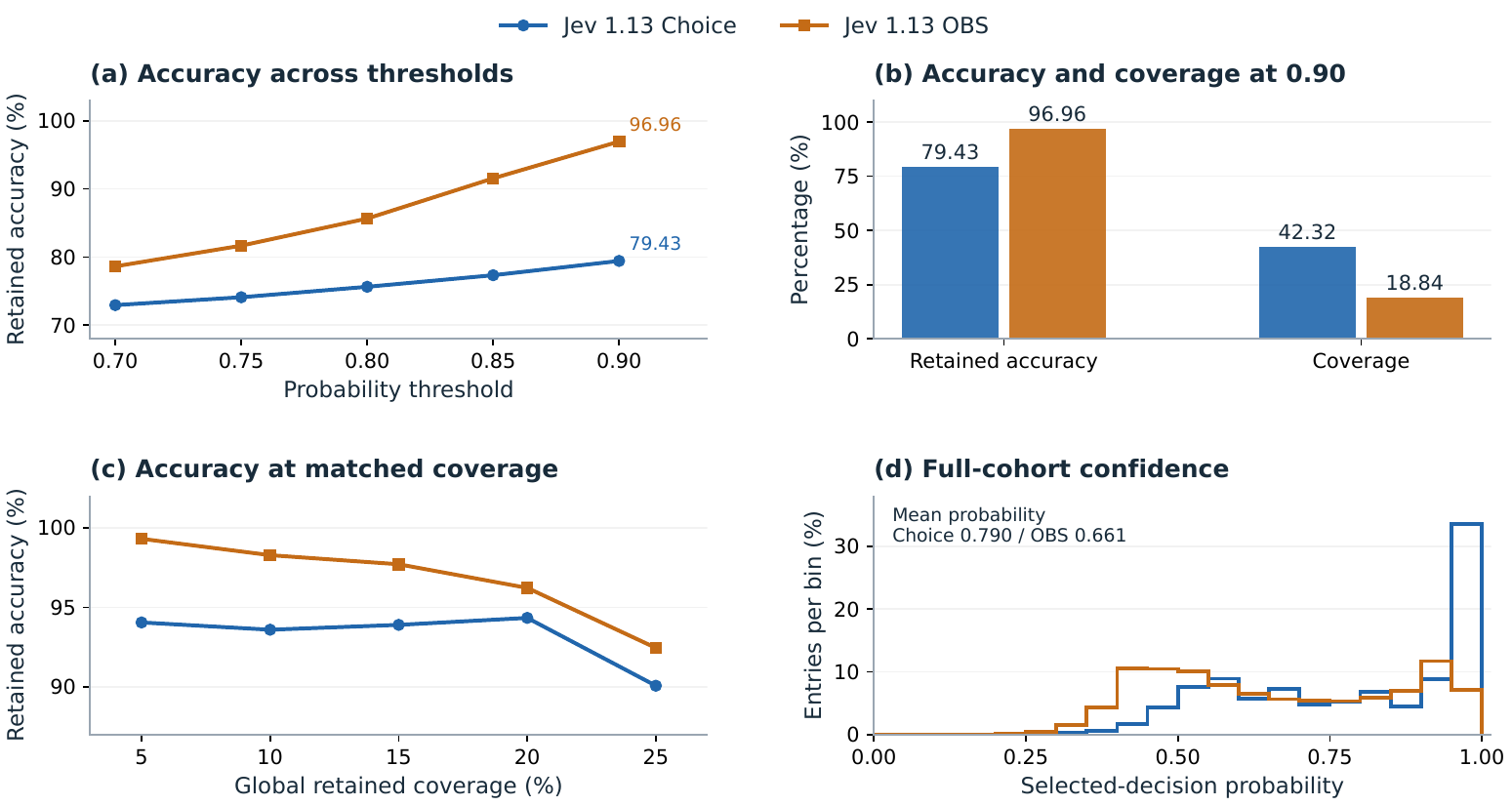}
\caption{\textbf{Jev 1.13's probability-based selection.} Choice (blue) and OBS (orange) compared through (a) threshold accuracy, (b) accuracy and coverage at 0.90, (c) accuracy at matched global coverage, and (d) confidence distributions (bin width 0.05). Accuracy pools retained entries; coverage uses all 8,742 entries. These are selection results, not calibration curves.}
\vspace{-0.3cm}
\label{fig:jev_probability}
\end{figure*}

\begin{figure*}[!t]
\centering
\includegraphics[width=0.88\textwidth]{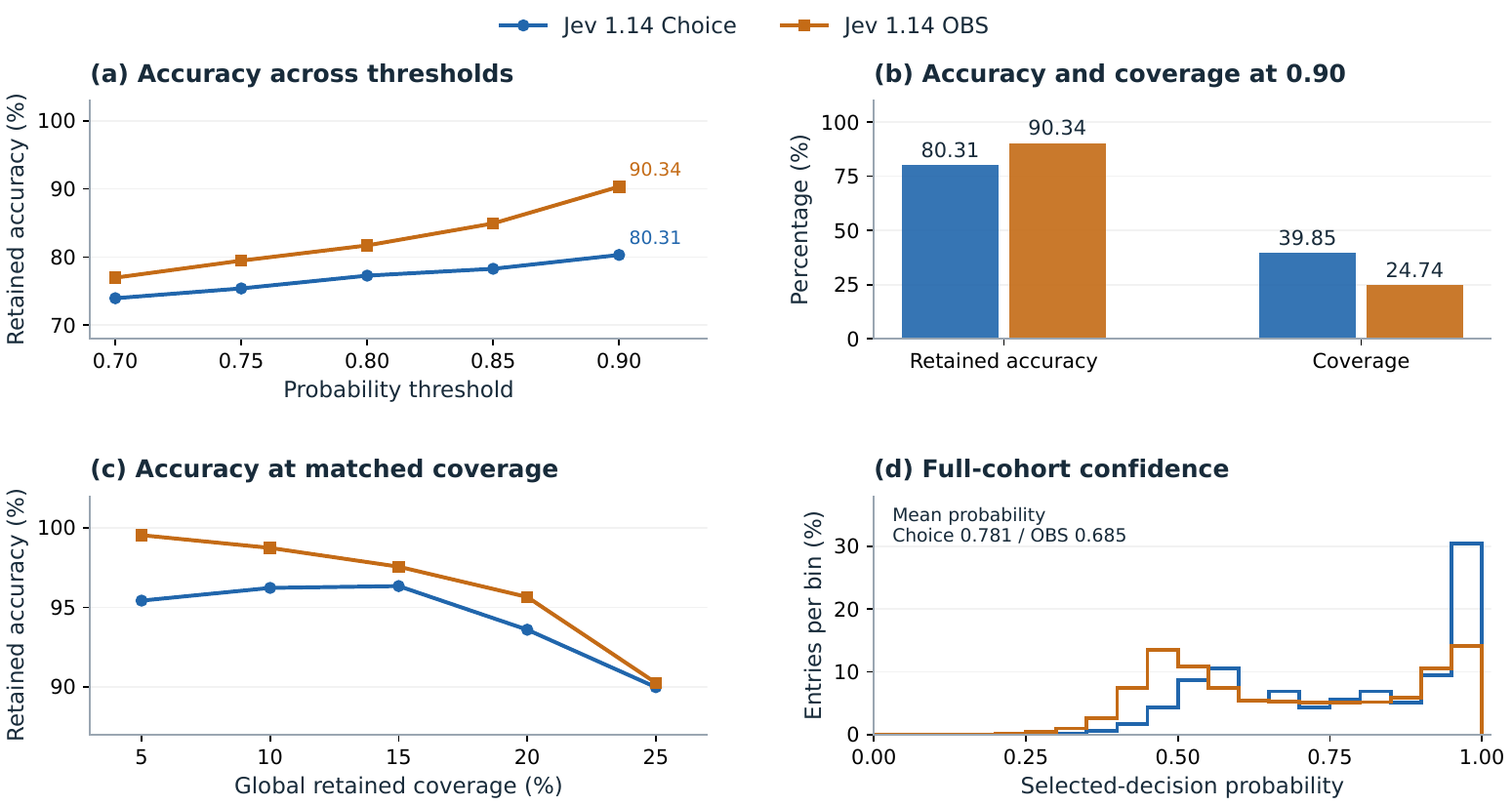}
\caption{\textbf{Jev 1.14's probability-based selection.} Choice (blue) and OBS (orange) compared through (a) threshold accuracy, (b) accuracy and coverage at 0.90, (c) accuracy at matched global coverage, and (d) confidence distributions (bin width 0.05). Accuracy pools retained entries; coverage uses all 8,742 entries. These are selection results, not calibration curves.}
\vspace{-0.3cm}
\label{fig:jev_probability_114}
\end{figure*}

Decision probabilities can guide downstream routing, indicating which answers to accept and which to defer for additional computation or review~\citep{selective,frugalgpt,routellm}. Useful signals must distinguish correct decisions from errors, not merely assign high scores. We compare direct Choice with normalized option-wise binary scoring (OBS) for Jev's probability-based answer selection.

\subsection{Evaluation Setup}
We pool all 8,742 entries and compare retained accuracy and coverage at probability thresholds $0.70/0.75/0.80/0.85/0.90$ and global top-$5/10/15/20/25\%$ selection. Confidence is the probability of the selected answer. Matched coverage fixes retained counts, not item identities. We also compare full-cohort probability distributions. Definitions, cross-model results, and task-level coverage appear in Appendices~\ref{app:confidence}--\ref{app:task_coverage}.

\subsection{Results and Analysis}

\paragraph{OBS improves threshold selection, but retains fewer decisions.}
Across both Jev versions, retained accuracy increases more steeply under OBS than under Choice as the threshold rises from $0.70$ to $0.90$ (Figures~\ref{fig:jev_probability}(a) and~\ref{fig:jev_probability_114}(a)). At $0.90$, OBS achieves $90.34$--$96.96\%$ retained accuracy, compared with $79.43$--$80.31\%$ under Choice, yielding gains of $10.03$--$17.53$ percentage points. These gains come with lower coverage: $18.84$--$24.74\%$ under OBS versus $39.85$--$42.32\%$ under Choice (Figures~\ref{fig:jev_probability}(b) and~\ref{fig:jev_probability_114}(b)). OBS is more accurate at every tested threshold, but a shared cutoff does not equate selection budgets.

\paragraph{The advantage persists at matched coverage.}
OBS also outperforms Choice at all five global retention levels in both versions (Figures~\ref{fig:jev_probability}(c) and~\ref{fig:jev_probability_114}(c)). At 5\% coverage, OBS achieves $99.31$--$99.54\%$ accuracy versus $94.05$--$95.42\%$ under Choice; at 25\%, the corresponding ranges are $90.25$--$92.45\%$ and $89.98$--$90.07\%$. With nearly unchanged unfiltered accuracy (Section~\ref{sec:robustness}), the main benefit is better probability-based answer selection.

\paragraph{Less extreme probabilities accompany better selection.}
OBS lowers mean selected-decision confidence in both Jev versions (Figures~\ref{fig:jev_probability}(d) and~\ref{fig:jev_probability_114}(d)). For Jev-1.13, Choice assigns probability one to 1,819 decisions, including 102 errors, whereas OBS produces no such scores. This illustrates reduced unwarranted certainty on some decisions; the evidence for better selection comes from retained accuracy rather than lower confidence alone.

\paragraph{High-confidence coverage aligns with task strengths.}
Jev's high-confidence decisions concentrate on tasks where it performs well, consistent with the task-specific strengths identified in Section~\ref{sec:performance}. At the $0.90$ OBS threshold, ICD-proxy and MIMIC-CDM, where both Jev versions exceed $94\%$ accuracy, account for $70.46$--$78.08\%$ of retained entries, while CTG-21 and MedXpertQA contribute none (Appendix~\ref{app:task_coverage}). Global selection gains therefore partly reflect concentration on stronger tasks, rather than uniform improvements across tasks. These benefits are model-dependent. Unlike Jev, D3-27B and all four StartLux configurations obtain lower retained accuracy under OBS at every tested matched coverage level. For D3-27B, accuracy at 20\% coverage decreases from $97.94\%$ under Choice to $91.53\%$ under OBS. The preferred scoring formulation must therefore be evaluated for each model (Appendices~\ref{app:confidence}--\ref{app:distributions}).




\section{Conclusion}
Across 23 model configurations and 15 medical datasets, accuracy alone does not establish reliable decisions. Jev combines competitive text-driven performance with strong format robustness, while D3-27B achieves higher overall scores and lower mean decision reversal. Neither resolves the limitations in abstention awareness. OBS improves Jev's probability-based selection but reduces retained accuracy for D3-27B at matched coverage, underscoring the need to evaluate task suitability, decision consistency, and scoring formulation jointly.

\clearpage


\section*{Limitations}

\paragraph{Coverage of an evolving model landscape.}
Growing interest in Jev has motivated an expanding range of Jev-like models with diverse designs. Given evaluation costs, our study focuses on representative configurations.
Its primary aim is to establish a common benchmark for assessing decision reliability in medicine. 
The reported comparisons capture a snapshot of the evaluated models; incorporating emerging alternatives under the same protocol will be important for tracking progress.

\paragraph{From isolated decisions to interactive environments.}
Our benchmark covers individual decisions across diverse medical tasks, but does not evaluate their role in interactive workflows. 
Jev could potentially support intermediate decisions in medical agents, such as selecting tools or determining the next action. 
Assessing these applications requires executable medical environments that capture changing information, sequential dependencies, and the consequences of earlier decisions. 
Our results do not establish Jev's effectiveness in agent-based settings. Further evaluation is needed to determine whether its strengths in isolated decisions translate into reliable end-to-end behavior.


\makeatletter
\global\@colroom\textheight
\global\@colht\textheight
\global\vsize\textheight
\makeatother
\bibliography{references,model_sources,routing_sources}
\clearpage
\onecolumn
\raggedbottom
\appendix
\nolinenumbers
\raggedcolumns
\setlength{\multicolsep}{6pt}

\begin{multicols}{2}
\section{Extended Related Work}
\label{app:related_extended}

\subsection{Generalizable Decision Models}

Jev formulates decision-making as instruction-conditioned candidate selection rather than unrestricted text generation, enabling direct decisions across heterogeneous tasks~\citep{typesafe}. Prior work has examined its zero-shot capabilities~\citep{jev_general}, scientific decision-making~\citep{jev_science}, and probability semantics~\citep{sys1cal}. Related models, including SemIf, Kev, AnyJev, GLiNER2.5-Decide, OpenJev, and StartLux, implement similar decision interfaces through different mechanisms~\citep{semif_repo,kev_repo,anyjev_repo,gliner_decide_card,openjev_repo,startlux_models}. D3-27B further provides a native interface for candidate selection and binary judgments.\footnote{\url{https://huggingface.co/vllm-sr/d3}} Studies of option naming and adversarial inputs further indicate that such decisions may remain sensitive to presentation changes~\citep{jev_names,jev_adv}.

Medical evaluations of Jev remain limited to a public examination pilot~\citep{korean_pilot}, a 2,823-item reliability study~\citep{jev_medicine}, and a radiology factuality assessment~\citep{jev_radiology}. Although informative, these studies cover relatively narrow settings and largely emphasize answer accuracy. How Jev and related models compare across diverse medical tasks, decision formats, and reliability dimensions therefore remains unclear.

\subsection{Medical Evaluation and Reliability}

Medical evaluation has expanded from knowledge examinations~\citep{medqa,singhal,medxpert} to benchmarks involving medical records, diagnostic and treatment decisions, agent-based tasks, and health-related conversations~\citep{medhelm,meds,hager,clibench,medagent,healthbench}. However, aggregate accuracy alone cannot establish whether decisions remain reliable under changes in evidence, candidate presentation, or answer availability.

Reliability requires complementary evaluation dimensions. Prior studies examine robustness to medical-case perturbations and answer presentation~\citep{medfuzz,order,selectionbias}, recognition of inadequate answer sets and appropriate abstention~\citep{know,metamed}, selective decision-making~\citep{selective}, and probability calibration~\citep{guo}. Motivated by these perspectives, our unified candidate-based framework evaluates Jev, Jev-like models, and LLMs in terms of \emph{format robustness}, \emph{abstention awareness}, and \emph{probability-based reliability}, in addition to overall performance.

\end{multicols}
\clearpage
\begin{multicols}{2}

\section{Task Composition and Adaptation Details}
\label{app:preliminary_details}
We define the medical task families, model groups, and shared decision protocol used throughout the evaluation.

\subsection{Medical Decision Tasks and Datasets}

Our evaluation comprises 8,742 entries from 15 selected or adapted medical datasets. Each entry presents a task, the available evidence, and a set of candidate answers, from which the model selects one response. This common formulation enables comparisons across models while preserving the distinct targets and references of the source tasks. Table~\ref{tab:benchmark_inventory} summarizes dataset sources, input evidence, decision targets, candidate counts, and evaluation sizes.

We group the tasks into four families according to the decision requested. These families provide complementary settings for examining the capabilities and limitations of direct decision models.

\paragraph{A: Triage and Navigation.}
These tasks concern when or where a patient should seek care. Models assign urgency or select a care-seeking action using patient disclosures, presenting complaints, and available clinical measurements.
This family includes CARE~\citep{care}, ESI scenarios drawn from MIMIC-IV and TriageAgent~\citep{mimiciv,triageagent}, a MIMIC-IV-ED cohort, and Ramaswamy's care-seeking scenarios~\citep{ramaswamy}.

\paragraph{B: Clinical Judgment.}
These tasks require interpreting clinical evidence to select a diagnosis, management action, coding category, or judgment about the effect of new evidence. Inputs combine clinical narratives, structured findings, and, where applicable, explicit scoring rules.
They include threshold judgments adapted from MedCalc~\citep{medcalc} and script-concordance judgments from SCT~\citep{sct_repo}, alongside diagnosis, coding, and examination tasks.

\paragraph{C: Safety Review.}
These tasks require assessing the correctness or safety of an existing clinical statement, document, or treatment candidate. Decisions depend on the supplied clinical context, source evidence, or patient-specific constraints.
We use MEDEC~\citep{medec} for error detection, MedVAL~\citep{medval} for document-risk assessment, and RxSafe~\citep{rxsafe} for medication selection under patient-specific safety constraints.

\paragraph{D: Physiological Assessment.}
These tasks use numerical observations as the principal evidence for assessing a physiological state or predicting a future event. They include static feature-based classification and short-horizon forecasting from time-series observations. Numerical measurements also occur in A--C; their central role in the decision distinguishes D.
The two tasks use CTG summary features~\citep{ctg_source} and intraoperative observations from VitalDB~\citep{vitaldb}, respectively.

We adapt calculation tasks into threshold judgments with supplied scoring rules and interactive tasks into static decisions based on available evidence. Reference answers follow the source datasets, including examination answers, recorded labels, expert judgments, and observation-derived outcomes. Each dataset contributes at most 1,000 entries; some cohorts share source material or underlying scenarios. Appendix~\ref{app:examples} provides examples, adaptations, and reference definitions.

\end{multicols}
\begin{table}[H]
\centering\small\setlength{\tabcolsep}{4pt}
\begin{tabular}{llrr>{\raggedright\arraybackslash}p{4.6cm}>{\raggedright\arraybackslash}p{5.7cm}}
\toprule
Domain & Dataset & $N$ & $K$ & Input evidence & Classification target \\
\midrule
A & CARE & 135 & 4 & Chronological patient messages & Clarify / self-care / nonurgent care / urgent care$^{\dagger}$\\
& ESI & 1,000 & 5 & Triage records and clinical vignettes & ESI 1 / ESI 2 / ESI 3 / ESI 4 / ESI 5$^{\dagger}$\\
& MIMIC-IV-ED & 1,000 & 5 & Chief complaint, demographics, pain, vital signs & ESI 1 / ESI 2 / ESI 3 / ESI 4 / ESI 5$^{\dagger}$\\
& Ramaswamy & 960 & 4 & Care-seeking narratives with available measurements & Home care / care within weeks / care within 24--48 hours / immediate ED$^{\dagger}$\\
& \textbf{A subtotal} & \textbf{3,095} & & & Triage and Navigation\\
\midrule
B & ICD-proxy & 1,000 & 4 & Admission-related narrative excerpts & ICD candidates A / B / C / D$^{\ddagger}$; e.g., D: J15 (bacterial pneumonia)\\
& MedCalc & 41 & 2 & Patient note, calculation question, scoring rules & Threshold not met / threshold met$^{\dagger}$\\
& MedQA & 252 & 4 & Clinical vignette and clinical question & Clinical candidates A / B / C / D$^{\ddagger}$; e.g., B: ketotifen eye drops\\
& MedXpertQA & 130 & 10 & Acute clinical vignette and clinical question & Clinical candidates A / B / $\cdots$ / J$^{\ddagger}$; e.g., B: identify the threatened person to police\\
& MIMIC-CDM & 1,000 & 4 & Clinical history, examination findings, reports, laboratory results & Appendicitis / cholecystitis / diverticulitis / pancreatitis$^{\dagger}$\\
& SCT & 78 & 5 & Clinical scenario, hypothesis, new evidence & Support change $-2$ / $-1$ / $0$ / $+1$ / $+2$$^{\dagger}$\\
& \textbf{B subtotal} & \textbf{2,501} & & & Clinical Judgment\\
\midrule
C & MEDEC & 509 & 2 & Clinical document under review & Error absent / error present$^{\dagger}$\\
& MedVAL & 388 & 4 & Source document, candidate document, document purpose & No risk / low risk / moderate risk / high risk$^{\dagger}$\\
& RxSafe & 915 & 3 & Consultation dialogue and medication candidates & Medication candidates A / B / C$^{\ddagger}$; e.g., A: Harvoni\\
& \textbf{C subtotal} & \textbf{1,812} & & & Safety Review\\
\midrule
D & CTG-21 & 1,000 & 3 & Numerical CTG summary features & Normal / suspect / pathologic$^{\dagger}$\\
& VitalDB & 334 & 2 & Time-stamped MAP observations, age, sex & Low-MAP event absent / low-MAP event present$^{\dagger}$\\
& \textbf{D subtotal} & \textbf{1,334} & & & Physiological Assessment\\
\midrule
& \textbf{Total} & \textbf{8,742} & & \multicolumn{2}{l}{\textbf{15 datasets; single-selection baseline questions.}}\\
\bottomrule
\end{tabular}
\caption{Benchmark composition and input evidence. Inputs include task instructions, candidate answers, and applicable definitions or constraints. $N$ denotes the number of entries, not independent patients; $K$ denotes the number of candidates in the baseline task. $\dagger$ indicates fixed class meanings within a dataset, while $\ddagger$ indicates question-specific candidates, with illustrative options shown in the table. Full task examples and adaptation details are provided in Appendix~\ref{app:examples}.}
\label{tab:benchmark_inventory}
\end{table}

\clearpage
\begin{multicols}{2}

\section{Dataset Inputs, Targets, and Examples}
\label{app:data}
\label{app:examples}

Table~\ref{tab:benchmark_inventory} summarizes the evaluated datasets, sample counts, and candidate classes. Tables~\ref{tab:examplesA}--\ref{tab:examplesD} present abridged, paraphrased examples of task inputs and source-derived reference answers. The examples are condensed for readability, while evaluation uses the complete stored inputs. Reference answers, including recorded ESI and coding labels, follow the source datasets and their task-specific definitions.

\end{multicols}
\begin{table}[H]
\centering\small\setlength{\tabcolsep}{5pt}
\begin{tabular}{p{2.15cm}p{9.1cm}p{3.55cm}}
\toprule
Dataset / target & Example input (abridged) & Expected benchmark output \\
\midrule
CARE / 4 actions & Ongoing root-canal treatment with severe tooth pain; the disclosed follow-up denies fever, facial swelling, pus, swallowing or breathing difficulty. Choose A: clarify; B: self-care; C: nonurgent care; D: urgent evaluation. & C: recommend nonurgent professional care.\\[5pt]
ESI / 5 levels & An eight-month-old with fever, cough, vomiting, reduced intake, listlessness, and a sunken fontanel; respirations are regular and unlabored. Apply the supplied ESI decision rules. A--E correspond to ESI 1--5. & B: ESI 2, from the TriageAgent scenario reference.\\[5pt]
MIMIC-IV-ED / 5 levels & Initial complaint: burn. Age 30, male, walk-in; temperature 98.2 F, pulse 75/min, respirations 14/min, oxygen saturation 100\%, BP 124/77 mmHg, pain 4/10. Predict the recorded ESI among 1--5. & E: recorded ESI 5. This is a historical nurse label, not a newly inferred clinical gold standard.\\[5pt]
Ramaswamy / 4 destinations & Two resolved episodes of unilateral arm numbness; currently normal neurologic examination and listed normal glucose. Options: A home; B physician in weeks; C physician within 24--48 hours; D immediate ED. & C or D is accepted in this source item; the model still selects one option.\\
\bottomrule
\end{tabular}
\caption{A: Triage and Navigation, 3,095 entries. Output labels preserve each example's original letter mapping. These tasks assess care urgency or routing, not a full differential diagnosis.}
\label{tab:examplesA}
\end{table}

\begin{table}[H]
\centering\small\setlength{\tabcolsep}{5pt}
\begin{tabular}{p{2.15cm}p{9.1cm}p{3.55cm}}
\toprule
Dataset / target & Example input (abridged) & Expected benchmark output \\
\midrule
ICD-proxy / 4 categories & Admission-related narrative describes cough, dyspnea, feverishness, and severe multifocal pneumonia. Candidates: A G40 epilepsy; B S02 skull/facial fracture; C S82 lower-leg fracture; D J15 bacterial pneumonia. & D: J15, the source's first-listed three-character ICD category.\\[5pt]
MedCalc / binary flag & Original question asks for admission GCS before treatment. Note describes no eye opening, incomprehensible vocalization, and abnormal flexion to pain. Supplied rules yield $1+2+3=6$; test whether GCS $\leq8$. A: flag absent; B: flag present. & B: threshold flag met. This does not itself specify airway treatment.\\[5pt]
MedQA / 4 alternatives & Recurrent springtime itchy, watery eyes and sneezing; bilateral conjunctival injection, normal visual acuity, and unremarkable anterior chamber. Best treatment? A erythromycin; B ketotifen; C warm compresses; D fluorometholone. & B: ketotifen eye drops, retaining the source examination answer.\\[5pt]
MedXpertQA / 10 alternatives & An oriented 19-year-old describes a specific plan to kill his father. A police officer requests the threatened person's identity. The source asks for the most appropriate next step among ten original responses, including B: identify the father to the officer. & B: the original examination reference. All ten alternatives are retained during evaluation.\\[5pt]
MIMIC-CDM / 4 diagnoses & Right upper quadrant and epigastric pain; focal tenderness; ultrasound describes gallbladder distension, stones/sludge, and wall thickening. The actual input also includes the remaining work-up. A appendicitis; B cholecystitis; C diverticulitis; D pancreatitis. & B: cholecystitis. This is a static diagnosis decision, not tool selection.\\[5pt]
SCT / 5 evidence updates & A febrile three-week-old infant is being evaluated. Hypothesis: starting acyclovir. New evidence: a brief seizure-like episode. Choose change in appropriateness from $-2$ to $+2$ (A--E). & E ($+2$) has maximum expert credit 1.00; D ($+1$) receives partial credit 0.25, not an accuracy hit here.\\
\bottomrule
\end{tabular}
\caption{B: Clinical Judgment, 2,501 entries. Different tasks retain examination answers, coding proxies, recorded diagnoses, explicit numerical rules, or expert-credit references; they are not interchangeable gold-standard constructions.}
\label{tab:examplesB}
\end{table}

\begin{table}[H]
\centering\small\setlength{\tabcolsep}{5pt}
\begin{tabular}{p{2.15cm}p{9.1cm}p{3.55cm}}
\toprule
Dataset / target & Example input (abridged) & Expected benchmark output \\
\midrule
MEDEC / error detection & Clinical note describes fever, diarrhea, abdominal pain and a rash after freshwater activities during travel to Brazil, and attributes the presentation to hepatitis A. Judge medical correctness: A no error; B an error is present. & B: medical error present, as labeled in the source. No localization or correction is required.\\[5pt]
MedVAL / 4 risk grades & Source dialogue discusses urinary symptoms, a prostate evaluation, laboratory tests, Flomax and specialist referral. The candidate note additionally asserts a suspected urinary infection and an antibiotic course. The complete original and candidate documents are supplied for review. A no risk; B low; C moderate; D high. & C: moderate document risk; expert review required, according to the physician-assigned source grade.\\[5pt]
RxSafe / 3 medication candidates & Consultation describes hepatitis C and concurrent atorvastatin use. Choose an appropriate medication while considering the described interaction risk. Candidates: A Harvoni; B Mavyret; C hyoscyamine. & A: Harvoni, the source reference among these candidates; not a general claim that the drug has no interaction risk.\\
\bottomrule
\end{tabular}
\caption{C: Safety Review, 1,812 entries. The reviewed object is a clinical assertion, generated document, or candidate medication. Document risk is distinct from illness severity; candidate selection is distinct from unrestricted prescribing.}
\label{tab:examplesC}
\end{table}

\begin{table}[H]
\centering\small\setlength{\tabcolsep}{5pt}
\begin{tabular}{p{2.15cm}p{9.1cm}p{3.55cm}}
\toprule
Dataset / target & Example input (abridged) & Expected benchmark output \\
\midrule
CTG-21 / 3 states & Fetal heart-rate baseline 140; AC 0; FM 0.00142; UC 0.00284; DL 0.00709; DS/DP 0; ASTV 62; MSTV 1.2; ALTV 0; MLTV 3.9; histogram width 93, minimum 66, maximum 159, peaks 5, zeros 2, mode 142, mean 136, median 144, variance 26, tendency 1. Definitions and source units accompany every field. & A: normal fetal state. Other candidates are B: suspect and C: pathologic. All 21 features, not a raw CTG image, are used.\\[5pt]
VitalDB / future event & An intraoperative case supplies age/sex and 60 pre-decision ten-second MAP medians, beginning $(-600,84),(-590,86),(-580,85),\ldots$ in seconds/mmHg. Predict whether the next five minutes contain six consecutive bins below 65 mmHg. A: no; B: yes. & A: no qualifying future episode in this item. The reference is derived from future measurements that are \emph{not} part of the input.\\
\bottomrule
\end{tabular}
\caption{D: Physiological Assessment, 1,334 entries. CTG values are rounded here for readability; evaluation preserves source precision. VitalDB is a probabilistic forecast with an observed future label, not a deterministically implied outcome from the abbreviated history.}
\label{tab:examplesD}
\end{table}

\clearpage
\begin{multicols}{2}

\section{Models and Input Protocol}
\label{app:models}

\end{multicols}
\begin{table}[H]
\centering\small
\setlength{\tabcolsep}{4pt}
\begin{tabular}{>{\raggedright\arraybackslash}p{1.35cm}>{\raggedright\arraybackslash}p{3.25cm}>{\raggedright\arraybackslash}p{4.15cm}>{\raggedright\arraybackslash}p{6.05cm}}
\toprule
Family & Evaluated configuration & Source & Decision readout \\
\midrule
Jev
& Jev-1.13\newline Jev-1.14
& TypeSafe~\citep{typesafe}
& C: native choice probabilities; B: native \texttt{noul} probabilities. \\
\midrule
Jev-like & GLiNER2.5-Decide & Fastino~\citep{gliner_decide_card} & C: candidate-logit softmax; B: two-class-logit softmax.\\
& SemIf & TheoLeeCJ~\citep{semif_repo} & C: candidate-token softmax; B: two-candidate-token softmax.\\
& Kev-4B & Jared Palmer~\citep{kev_card} & C: native decision-head probabilities; B: native \texttt{noul} probabilities.\\
& AnyJev (raw) & Nokia Applied Research~\citep{anyjev_repo} & C: candidate-slot softmax; B: native yes/no probabilities.\\
& OpenJev & OpenJev~\citep{openjev_repo} & C: answer-slot softmax; B: binary answer-slot softmax.\\
& StartLux-0.8B; 4B; 9B; 27B & StartLux Models~\citep{startlux_models} & C: native choice probabilities; B: native \texttt{noul} probabilities.\\
 & D3-27B & vLLM Semantic Router (\href{https://huggingface.co/vllm-sr/d3}{model card}) & C: native choice probabilities; B: native \texttt{noul} probabilities. \\
\midrule
\multicolumn{3}{@{}l@{}}{%
\begin{tabular}[c]{>{\raggedright\arraybackslash}p{1.35cm}>{\raggedright\arraybackslash}p{3.25cm}>{\raggedright\arraybackslash}p{4.15cm}}
LLM: general & Qwen3.5-0.8B; 4B; 9B & Qwen~\citep{qwen35_08_card,qwen35_4_card,qwen35_9_card} \\
& Qwen3.6-27B; Qwen3.8-27B & Qwen~\citep{qwen36_card,qwen38_card} \\
& Gemma4-12B-IT; 31B-IT & Google~\citep{gemma4_12_card,gemma4_31_card} \\
& Kimi-K2.6 & Moonshot AI~\citep{kimi26_card} \\
\cmidrule(lr){1-3}
LLM: medical & MedGemma-4B-IT; 27B-IT & Google~\citep{medgemma4_card,medgemma27_card} \\
& Baichuan-M3-235B & Baichuan~\citep{baichuan3_card}
\end{tabular}} & \parbox[c]{6.05cm}{\raggedright\textbf{Shared by all eleven LLMs:}\par C: candidate-label-token softmax;\par B: binary-label-token softmax.\par Both use the same fixed-prefix readout under the system-one protocol.}\\
\bottomrule
\end{tabular}

\caption{Model provenance and evaluated readouts. Multi-size rows identify separate configurations, totaling 23 models. C denotes the joint Choice distribution; B denotes the binary distribution for one candidate judgment. OBS extracts B's affirmative probability for each candidate and normalizes across candidates. All LLMs use the same fixed-prefix, single-token-label protocol with model-specific chat templates. Backbone and adapter details appear in Appendix~\ref{app:models}.}
\label{tab:model_inventory}
\end{table}
\begin{multicols}{2}

\paragraph{Candidate probabilities.}
For LLM label log-probabilities $\ell_k$, Choice uses
\begin{equation}
p_k=\frac{\exp(\ell_k)}{\sum_{j=1}^{K}\exp(\ell_j)}.
\label{eq:choice}
\end{equation}
OBS normalizes the affirmative probability $q_k$ of each candidate judgment as
\begin{equation}
\widetilde p_k=\frac{q_k}{\sum_{j=1}^{K}q_j}.
\label{eq:preliminary_obs}
\end{equation}
Both select the maximum-scoring candidate. Binary judgments retain the complete task context and candidate set.

Table~\ref{tab:model_inventory} lists the evaluated configurations and
their sources. Jev refers to the model family; the evaluated service versions are Jev-1.13 and Jev-1.14~\citep{typesafe,typesafe_limits}. The Jev-like group comprises
models with related decision interfaces and diverse architectures and
implementations. The four StartLux sizes are evaluated as separate
configurations within the same model family. The following details
specify the evaluated backbones, adapters, and inference settings.

\paragraph{Unified Jev inputs and native interfaces.}
SemIf and Kev use Qwen3.5-4B-Base; Kev adds the released LoRA,
pointer head, and temperature. AnyJev uses Qwen3-4B through its raw
interface without a fitted medical head. GLiNER2.5-Decide uses
DeBERTa-v3-large. OpenJev's backbone is specified below.
All ten Jev-like configurations receive complete Jev-format requests,
whereas LLMs receive the corresponding chat-formatted inputs.
The adapters preserve the state, question instructions, types, criteria,
and candidate order; gold references remain outside the request.
StartLux uses the released \texttt{StartLuxDecision.decide} interface
and checkpoint temperatures. Kev uses its native request representation
and unrounded decision-head probabilities. Both support native Choice
and \texttt{noul} judgments.

\paragraph{D3 implementation.}

D3-27B receives the complete Jev-format state and questions through its native \texttt{system\_one} interface. Choice probabilities are mapped to the supplied candidate identifiers. For OBS, each native \texttt{noul} judgment retains both criterion descriptions, and its affirmative probability is normalized across candidates. No LLM-style answer prefill or elicited reasoning is used.
\paragraph{Content-preserving text adapters.}
AnyJev serializes the original state and the full question JSON into its
raw interface, with original candidate identifiers and descriptions.
Its Choice readout uses internal letter slots, and its native binary
readout maps affirmative/negative answers to the supplied criteria.
SemIf receives the serialized state as evidence and the complete question
JSON as its criterion, using fixed internal letter slots for candidate
scoring. GLiNER receives the complete state/question JSON and the
candidate descriptions as classification inputs, with a softmax over
class logits. SemIf and GLiNER implement OBS as content-preserving
two-class adapters, not native \texttt{noul} heads. Fixed internal
slots do not remove displayed identifiers from the full question text;
relabeling can therefore still affect these adapters' predictions.
The reported robustness is that of each evaluated model--adapter pair.

\paragraph{OpenJev implementation.}
OpenJev uses diffusiongemma-26B-A4B-it with the native prompt and canvas
construction, BF16 eager inference, one diffusion step, one sample,
and seed 1234 for each request. Candidate probabilities are softmax
logits at the answer slot.

\paragraph{LLM readout and numeric labels.}
LLMs use their model-specific chat templates without elicited reasoning.
The evaluator forcibly appends \texttt{The answer is \textbackslash boxed\{}
as an assistant-response prefill, rather than asking the model to
generate this prefix. A forward pass reads the next-token distribution
immediately after the opening brace. The log-probabilities of valid
single-token candidate labels are softmax-normalized over those labels
only, equivalently renormalizing their token probabilities. The label
with the largest normalized probability determines the prediction;
neither sampled continuations nor answer-sequence likelihoods are used.
OBS applies the identical prefill to each candidate proposition,
normalizes over its affirmative and negative label tokens, and then
renormalizes the affirmative probabilities across candidates.

All binary source tasks are represented as two-candidate choices in the baseline. Under \obs{}, each candidate is evaluated through a separate binary judgment, using A=True and B=False as the response labels. An illustrative, synthetic prompt is: ``Given the patient evidence and the complete candidate set, is candidate [ID: meaning] the correct response under the stated task rule?'' Each request retains the dataset-specific instructions, constraints, field definitions, and observations. Clinical ``no event'' labels describe task outcomes, binary ``false'' labels indicate rejection of a candidate proposition, and \texttt{ABSTAIN} denotes optional withholding.

The abstention analysis uses direct choice and is restricted to B: Clinical Judgment. Answer removal uses 2,382 entries from ICD-proxy, MedQA, MedXpertQA, and MIMIC-CDM; optional abstention uses all six B datasets (2,501 entries). SCT and all $K\leq3$ tasks are excluded from answer removal.

\end{multicols}
\clearpage
\begin{multicols}{2}

\section{Overall Performance: Full Accuracy and Macro-F1}
\label{app:performance}

\end{multicols}
\begin{figure}[H]
\centering\includegraphics[width=0.985\textwidth]{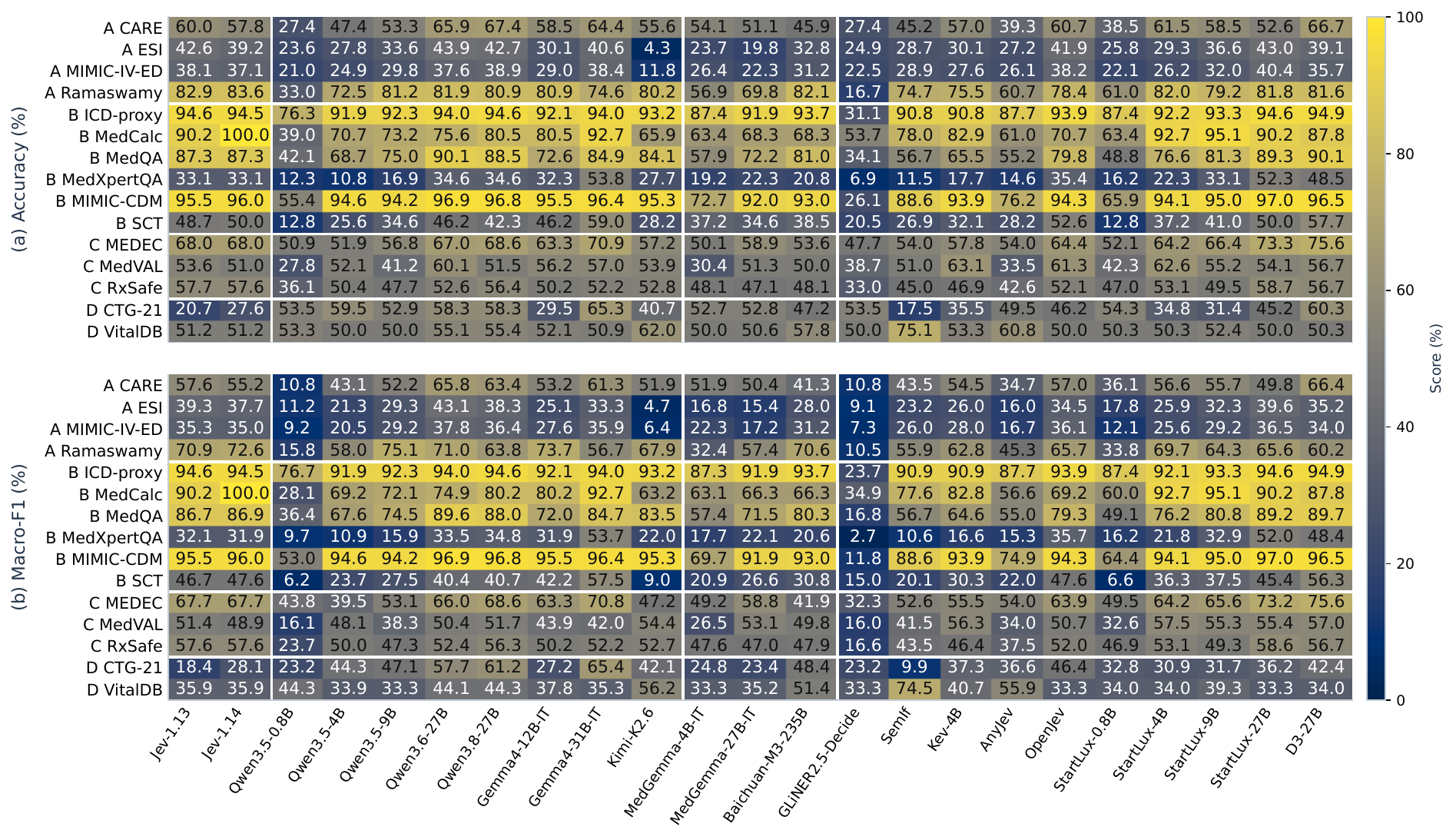}
\caption{Dataset-level original-choice \textbf{accuracy (top) and macro-F1 (bottom)} (\%) for all 23 model configurations.}
\label{fig:accuracy}
\end{figure}

\begin{multicols}{2}

Figure~\ref{fig:accuracy} presents the annotated accuracy and macro-F1 heatmaps, while Tables~\ref{tab:baseline1}--\ref{tab:baseline3} provide the corresponding dataset-level results for all models.Macro-F1 is computed on entries with unique references using the original candidate inventory. For examination-style tasks, candidate classes represent answer identities, with macro-F1 reflecting performance across these answer options.

\end{multicols}
\begin{table}[H]
\centering\footnotesize
\setlength{\tabcolsep}{2pt}
\begin{tabular*}{\textwidth}{@{\extracolsep{\fill}}lrrrrrr@{}}
\toprule
 & \multicolumn{2}{c}{\textit{Jev}} & \multicolumn{4}{c}{\textit{General-purpose LLMs}} \\
\cmidrule(lr){2-3}\cmidrule(lr){4-7}
Dataset & \shortstack{Jev-\\1.13} & \shortstack{Jev-\\1.14} & \shortstack{Qwen3.5-\\0.8B} & \shortstack{Qwen3.5-\\4B} & \shortstack{Qwen3.5-\\9B} & \shortstack{Qwen3.6-\\27B} \\
\midrule
A CARE & 60.0 / 57.6 & 57.8 / 55.2 & 27.4 / 10.8 & 47.4 / 43.1 & 53.3 / 52.2 & 65.9 / 65.8 \\
A ESI & 42.6 / 39.3 & 39.2 / 37.7 & 23.6 / 11.2 & 27.8 / 21.3 & 33.6 / 29.3 & 43.9 / 43.1 \\
A MIMIC-IV-ED & 38.1 / 35.3 & 37.1 / 35.0 & 21.0 / 9.2 & 24.9 / 20.5 & 29.8 / 29.2 & 37.6 / 37.8 \\
A Ramaswamy & 82.9 / 70.9 & 83.6 / 72.6 & 33.0 / 15.8 & 72.5 / 58.0 & 81.2 / 75.1 & 81.9 / 71.0 \\
B ICD-proxy & 94.6 / 94.6 & 94.5 / 94.5 & 76.3 / 76.7 & 91.9 / 91.9 & 92.3 / 92.3 & 94.0 / 94.0 \\
B MedCalc & 90.2 / 90.2 & 100.0 / 100.0 & 39.0 / 28.1 & 70.7 / 69.2 & 73.2 / 72.1 & 75.6 / 74.9 \\
B MedQA & 87.3 / 86.7 & 87.3 / 86.9 & 42.1 / 36.4 & 68.7 / 67.6 & 75.0 / 74.5 & 90.1 / 89.6 \\
B MedXpertQA & 33.1 / 32.1 & 33.1 / 31.9 & 12.3 / 9.7 & 10.8 / 10.9 & 16.9 / 15.9 & 34.6 / 33.5 \\
B MIMIC-CDM & 95.5 / 95.5 & 96.0 / 96.0 & 55.4 / 53.0 & 94.6 / 94.6 & 94.2 / 94.2 & 96.9 / 96.9 \\
B SCT & 48.7 / 46.7 & 50.0 / 47.6 & 12.8 / 6.2 & 25.6 / 23.7 & 34.6 / 27.5 & 46.2 / 40.4 \\
C MEDEC & 68.0 / 67.7 & 68.0 / 67.7 & 50.9 / 43.8 & 51.9 / 39.5 & 56.8 / 53.1 & 67.0 / 66.0 \\
C MedVAL & 53.6 / 51.4 & 51.0 / 48.9 & 27.8 / 16.1 & 52.1 / 48.1 & 41.2 / 38.3 & 60.1 / 50.4 \\
C RxSafe & 57.7 / 57.6 & 57.6 / 57.6 & 36.1 / 23.7 & 50.4 / 50.0 & 47.7 / 47.3 & 52.6 / 52.4 \\
D CTG-21 & 20.7 / 18.4 & 27.6 / 28.1 & 53.5 / 23.2 & 59.5 / 44.3 & 52.9 / 47.1 & 58.3 / 57.7 \\
D VitalDB & 51.2 / 35.9 & 51.2 / 35.9 & 53.3 / 44.3 & 50.0 / 33.9 & 50.0 / 33.3 & 55.1 / 44.1 \\
\bottomrule
\end{tabular*}
\par\bigskip
\begin{tabular*}{\textwidth}{@{\extracolsep{\fill}}lrrrr@{}}
\toprule
 & \multicolumn{4}{c}{\textit{General-purpose LLMs}} \\
\cmidrule(lr){2-5}
Dataset & \shortstack{Qwen3.8-\\27B} & \shortstack{Gemma4-\\12B-IT} & \shortstack{Gemma4-\\31B-IT} & Kimi-K2.6 \\
\midrule
A CARE & 67.4 / 63.4 & 58.5 / 53.2 & 64.4 / 61.3 & 55.6 / 51.9 \\
A ESI & 42.7 / 38.3 & 30.1 / 25.1 & 40.6 / 33.3 & 4.3 / 4.7 \\
A MIMIC-IV-ED & 38.9 / 36.4 & 29.0 / 27.6 & 38.4 / 35.9 & 11.8 / 6.4 \\
A Ramaswamy & 80.9 / 63.8 & 80.9 / 73.7 & 74.6 / 56.7 & 80.2 / 67.9 \\
B ICD-proxy & 94.6 / 94.6 & 92.1 / 92.1 & 94.0 / 94.0 & 93.2 / 93.2 \\
B MedCalc & 80.5 / 80.2 & 80.5 / 80.2 & 92.7 / 92.7 & 65.9 / 63.2 \\
B MedQA & 88.5 / 88.0 & 72.6 / 72.0 & 84.9 / 84.7 & 84.1 / 83.5 \\
B MedXpertQA & 34.6 / 34.8 & 32.3 / 31.9 & 53.8 / 53.7 & 27.7 / 22.0 \\
B MIMIC-CDM & 96.8 / 96.8 & 95.5 / 95.5 & 96.4 / 96.4 & 95.3 / 95.3 \\
B SCT & 42.3 / 40.7 & 46.2 / 42.2 & 59.0 / 57.5 & 28.2 / 9.0 \\
C MEDEC & 68.6 / 68.6 & 63.3 / 63.3 & 70.9 / 70.8 & 57.2 / 47.2 \\
C MedVAL & 51.5 / 51.7 & 56.2 / 43.9 & 57.0 / 42.0 & 53.9 / 54.4 \\
C RxSafe & 56.4 / 56.3 & 50.2 / 50.2 & 52.2 / 52.2 & 52.8 / 52.7 \\
D CTG-21 & 58.3 / 61.2 & 29.5 / 27.2 & 65.3 / 65.4 & 40.7 / 42.1 \\
D VitalDB & 55.4 / 44.3 & 52.1 / 37.8 & 50.9 / 35.3 & 62.0 / 56.2 \\
\bottomrule
\end{tabular*}

\caption{Original choice: Acc / macro-F1 (\%), panel 1 of 3. Model-family headings follow Table~\ref{tab:overall}. All 15 datasets are retained. A--D identify the four decision families. Ramaswamy F1 uses 480 unique-reference items and SCT F1 uses 76; accuracy uses all respective items.}
\label{tab:baseline1}
\label{tab:allbaseline}
\end{table}
\clearpage
\begin{table}[H]
\centering\footnotesize
\setlength{\tabcolsep}{2pt}
\begin{tabular*}{\textwidth}{@{\extracolsep{\fill}}lrrrr@{}}
\toprule
 & \multicolumn{3}{c}{\textit{Medical LLMs}} & \multicolumn{1}{c}{\textit{Jev-like models}} \\
\cmidrule(lr){2-4}\cmidrule(lr){5-5}
Dataset & \shortstack{MedGemma-\\4B-IT} & \shortstack{MedGemma-\\27B-IT} & \shortstack{Baichuan-\\M3-235B} & \shortstack{GLiNER2.5-\\Decide} \\
\midrule
A CARE & 54.1 / 51.9 & 51.1 / 50.4 & 45.9 / 41.3 & 27.4 / 10.8 \\
A ESI & 23.7 / 16.8 & 19.8 / 15.4 & 32.8 / 28.0 & 24.9 / 9.1 \\
A MIMIC-IV-ED & 26.4 / 22.3 & 22.3 / 17.2 & 31.2 / 31.2 & 22.5 / 7.3 \\
A Ramaswamy & 56.9 / 32.4 & 69.8 / 57.4 & 82.1 / 70.6 & 16.7 / 10.5 \\
B ICD-proxy & 87.4 / 87.3 & 91.9 / 91.9 & 93.7 / 93.7 & 31.1 / 23.7 \\
B MedCalc & 63.4 / 63.1 & 68.3 / 66.3 & 68.3 / 66.3 & 53.7 / 34.9 \\
B MedQA & 57.9 / 57.4 & 72.2 / 71.5 & 81.0 / 80.3 & 34.1 / 16.8 \\
B MedXpertQA & 19.2 / 17.7 & 22.3 / 22.1 & 20.8 / 20.6 & 6.9 / 2.7 \\
B MIMIC-CDM & 72.7 / 69.7 & 92.0 / 91.9 & 93.0 / 93.0 & 26.1 / 11.8 \\
B SCT & 37.2 / 20.9 & 34.6 / 26.6 & 38.5 / 30.8 & 20.5 / 15.0 \\
C MEDEC & 50.1 / 49.2 & 58.9 / 58.8 & 53.6 / 41.9 & 47.7 / 32.3 \\
C MedVAL & 30.4 / 26.5 & 51.3 / 53.1 & 50.0 / 49.8 & 38.7 / 16.0 \\
C RxSafe & 48.1 / 47.6 & 47.1 / 47.0 & 48.1 / 47.9 & 33.0 / 16.6 \\
D CTG-21 & 52.7 / 24.8 & 52.8 / 23.4 & 47.2 / 48.4 & 53.5 / 23.2 \\
D VitalDB & 50.0 / 33.3 & 50.6 / 35.2 & 57.8 / 51.4 & 50.0 / 33.3 \\
\bottomrule
\end{tabular*}
\par\bigskip
\begin{tabular*}{\textwidth}{@{\extracolsep{\fill}}lrrrr@{}}
\toprule
 & \multicolumn{4}{c}{\textit{Jev-like models}} \\
\cmidrule(lr){2-5}
Dataset & SemIf & Kev-4B & AnyJev & OpenJev \\
\midrule
A CARE & 45.2 / 43.5 & 57.0 / 54.5 & 39.3 / 34.7 & 60.7 / 57.0 \\
A ESI & 28.7 / 23.2 & 30.1 / 26.0 & 27.2 / 16.0 & 41.9 / 34.5 \\
A MIMIC-IV-ED & 28.9 / 26.0 & 27.6 / 28.0 & 26.1 / 16.7 & 38.2 / 36.1 \\
A Ramaswamy & 74.7 / 55.9 & 75.5 / 62.8 & 60.7 / 45.3 & 78.4 / 65.7 \\
B ICD-proxy & 90.8 / 90.9 & 90.8 / 90.9 & 87.7 / 87.7 & 93.9 / 93.9 \\
B MedCalc & 78.0 / 77.6 & 82.9 / 82.8 & 61.0 / 56.6 & 70.7 / 69.2 \\
B MedQA & 56.7 / 56.7 & 65.5 / 64.6 & 55.2 / 55.0 & 79.8 / 79.3 \\
B MedXpertQA & 11.5 / 10.6 & 17.7 / 16.6 & 14.6 / 15.3 & 35.4 / 35.7 \\
B MIMIC-CDM & 88.6 / 88.6 & 93.9 / 93.9 & 76.2 / 74.9 & 94.3 / 94.3 \\
B SCT & 26.9 / 20.1 & 32.1 / 30.3 & 28.2 / 22.0 & 52.6 / 47.6 \\
C MEDEC & 54.0 / 52.6 & 57.8 / 55.5 & 54.0 / 54.0 & 64.4 / 63.9 \\
C MedVAL & 51.0 / 41.5 & 63.1 / 56.3 & 33.5 / 34.0 & 61.3 / 50.7 \\
C RxSafe & 45.0 / 43.5 & 46.9 / 46.4 & 42.6 / 37.5 & 52.1 / 52.0 \\
D CTG-21 & 17.5 / 9.9 & 35.5 / 37.3 & 49.5 / 36.6 & 46.2 / 46.4 \\
D VitalDB & 75.1 / 74.5 & 53.3 / 40.7 & 60.8 / 55.9 & 50.0 / 33.3 \\
\bottomrule
\end{tabular*}

\caption{Original choice: Acc / macro-F1 (\%), panel 2 of 3, under the same scoring convention and model-family organization as Table~\ref{tab:baseline1}.}
\label{tab:baseline2}
\end{table}
\clearpage
\begin{table}[H]
\centering\small\setlength{\tabcolsep}{5pt}
\begin{tabular*}{\textwidth}{@{\extracolsep{\fill}}lrrrrr@{}}
\toprule
 & \multicolumn{5}{c}{\textit{Jev-like models}} \\
\cmidrule(lr){2-6}
Dataset & StartLux-0.8B & StartLux-4B & StartLux-9B & StartLux-27B & D3-27B \\
\midrule
A CARE & 38.5 / 36.1 & 61.5 / 56.6 & 58.5 / 55.7 & 52.6 / 49.8 & 66.7 / 66.4 \\
A ESI & 25.8 / 17.8 & 29.3 / 25.9 & 36.6 / 32.3 & 43.0 / 39.6 & 39.1 / 35.2 \\
A MIMIC-IV-ED & 22.1 / 12.1 & 26.2 / 25.6 & 32.0 / 29.2 & 40.4 / 36.5 & 35.7 / 34.0 \\
A Ramaswamy & 61.0 / 33.8 & 82.0 / 69.7 & 79.2 / 64.3 & 81.8 / 65.6 & 81.6 / 60.2 \\
B ICD-proxy & 87.4 / 87.4 & 92.2 / 92.1 & 93.3 / 93.3 & 94.6 / 94.6 & 94.9 / 94.9 \\
B MedCalc & 63.4 / 60.0 & 92.7 / 92.7 & 95.1 / 95.1 & 90.2 / 90.2 & 87.8 / 87.8 \\
B MedQA & 48.8 / 49.1 & 76.6 / 76.2 & 81.3 / 80.8 & 89.3 / 89.2 & 90.1 / 89.7 \\
B MedXpertQA & 16.2 / 16.2 & 22.3 / 21.8 & 33.1 / 32.9 & 52.3 / 52.0 & 48.5 / 48.4 \\
B MIMIC-CDM & 65.9 / 64.4 & 94.1 / 94.1 & 95.0 / 95.0 & 97.0 / 97.0 & 96.5 / 96.5 \\
B SCT & 12.8 / 6.6 & 37.2 / 36.3 & 41.0 / 37.5 & 50.0 / 45.4 & 57.7 / 56.3 \\
C MEDEC & 52.1 / 49.5 & 64.2 / 64.2 & 66.4 / 65.6 & 73.3 / 73.2 & 75.6 / 75.6 \\
C MedVAL & 42.3 / 32.6 & 62.6 / 57.5 & 55.2 / 55.3 & 54.1 / 55.4 & 56.7 / 57.0 \\
C RxSafe & 47.0 / 46.9 & 53.1 / 53.1 & 49.5 / 49.3 & 58.7 / 58.6 & 56.7 / 56.7 \\
D CTG-21 & 54.3 / 32.8 & 34.8 / 30.9 & 31.4 / 31.7 & 45.2 / 36.2 & 60.3 / 42.4 \\
D VitalDB & 50.3 / 34.0 & 50.3 / 34.0 & 52.4 / 39.3 & 50.0 / 33.3 & 50.3 / 34.0 \\
\bottomrule
\end{tabular*}

\caption{Original choice: Acc / macro-F1 (\%), panel 3 of 3, completing the Jev-like model block with the four StartLux configurations and D3-27B. Scoring and model-family organization follow Table~\ref{tab:baseline1}.}
\label{tab:baseline3}
\end{table}

\clearpage
\begin{multicols}{2}

\section{Format Robustness: Decision and Probability Consistency}
\label{app:format_scores}

\end{multicols}
\begin{table}[H]
\centering
\small
\renewcommand{\arraystretch}{0.98}
\setlength{\tabcolsep}{4pt}
\begin{tabular*}{\textwidth}{@{\extracolsep{\fill}}lrrrrrrrr@{}}
\toprule

Model & $\Delta$ Rel. & $\Delta$ Reord. & $\Delta$ OBS
& $V_{\Delta}$ & $F_{\mathrm{Rel}}$ & $F_{\mathrm{Reo}}$
& $F_{\mathrm{OBS}}$ & $\overline F$ \\

\midrule
\multicolumn{9}{@{}l}{\textit{Jev}} \\
Jev-1.13 & +0.45 & -0.03 & +0.05 & 1.70 & 3.9 & 6.3 & 8.3 & 6.2 \\
Jev-1.14 & +0.19 & +0.03 & +0.19 & 1.09 & 3.4 & 6.5 & 7.1 & 5.7 \\
\midrule
\multicolumn{9}{@{}l}{\textit{General-purpose LLMs}} \\
Qwen3.5-0.8B & -0.87 & +1.05 & -3.31 & 50.95 & 20.6 & 45.3 & 44.2 & 36.7 \\
Qwen3.5-4B & +0.28 & -0.05 & -4.63 & 21.51 & 10.7 & 18.9 & 40.9 & 23.5 \\
Qwen3.5-9B & +0.66 & +1.11 & -4.75 & 14.30 & 9.7 & 18.5 & 36.8 & 21.7 \\
Qwen3.6-27B & -0.17 & -0.34 & -3.55 & 18.29 & 5.4 & 9.2 & 16.1 & 10.2 \\
Qwen3.8-27B & +0.40 & -0.20 & -3.84 & 19.79 & 5.1 & 10.3 & 16.9 & 10.7 \\
Gemma4-12B-IT & +0.49 & -0.27 & -3.17 & 13.65 & 6.8 & 11.8 & 23.8 & 14.1 \\
Gemma4-31B-IT & +0.42 & -0.59 & -2.84 & 22.23 & 3.5 & 6.3 & 12.4 & 7.4 \\
Kimi-K2.6 & +6.51 & +1.94 & +5.42 & 44.32 & 23.2 & 27.2 & 31.6 & 27.3 \\
\midrule
\multicolumn{9}{@{}l}{\textit{Medical LLMs}} \\
MedGemma-4B-IT & -0.19 & -2.12 & -14.71 & 98.69 & 11.1 & 33.9 & 67.7 & 37.6 \\
MedGemma-27B-IT & +0.76 & +0.57 & -4.83 & 14.01 & 8.2 & 15.8 & 39.6 & 21.2 \\
Baichuan-M3-235B & +0.74 & +1.38 & -6.16 & 21.04 & 6.4 & 16.6 & 33.9 & 19.0 \\
\midrule
\multicolumn{9}{@{}l}{\textit{Jev-like models}} \\
GLiNER2.5-Decide & +0.31 & -2.03 & -1.02 & 20.76 & 16.1 & 71.7 & 23.0 & 36.9 \\
SemIf & -2.43 & -1.96 & -10.04 & 74.08 & 13.4 & 32.5 & 60.4 & 35.4 \\
Kev-4B & +0.58 & -1.02 & +0.12 & 2.42 & 4.9 & 13.4 & 15.9 & 11.4 \\
AnyJev & -0.68 & +0.31 & -5.97 & 14.23 & 15.0 & 37.4 & 44.5 & 32.3 \\
OpenJev & -0.68 & +0.66 & -2.73 & 6.85 & 8.2 & 13.6 & 19.3 & 13.7 \\
StartLux-0.8B & -0.31 & +0.11 & -1.76 & 5.26 & 9.4 & 11.5 & 18.3 & 13.1 \\
StartLux-4B & -0.27 & -0.75 & -0.39 & 1.26 & 4.8 & 8.6 & 9.5 & 7.6 \\
StartLux-9B & -0.23 & +0.38 & -0.28 & 1.31 & 4.4 & 8.1 & 9.2 & 7.2 \\
StartLux-27B & -0.08 & -0.14 & -0.25 & 6.45 & 2.9 & 6.5 & 10.0 & 6.5 \\
D3-27B & -0.24 & +0.33 & -0.41 & 2.62 & 2.5 & 4.5 & 7.3 & 4.7 \\
\bottomrule
\end{tabular*}

\caption{Format robustness across all 15 datasets, with equal
dataset weights. $\Delta$ Rel., $\Delta$ Reord., and $\Delta$ OBS
denote accuracy changes (percentage points) under relabeling, reordering, and OBS, respectively.
$V_{\Delta}$ denotes mean within-dataset accuracy variance
(squared percentage points).
$F_{\mathrm{Rel}}$, $F_{\mathrm{Reo}}$, and $F_{\mathrm{OBS}}$
denote prediction reversal rates (\%) under the corresponding
interventions and $\overline F$ is their mean.}
\label{tab:robustness_full}

\end{table}
\begin{multicols}{2}

\begin{figure}[H]
\centering
\includegraphics[width=\textwidth]{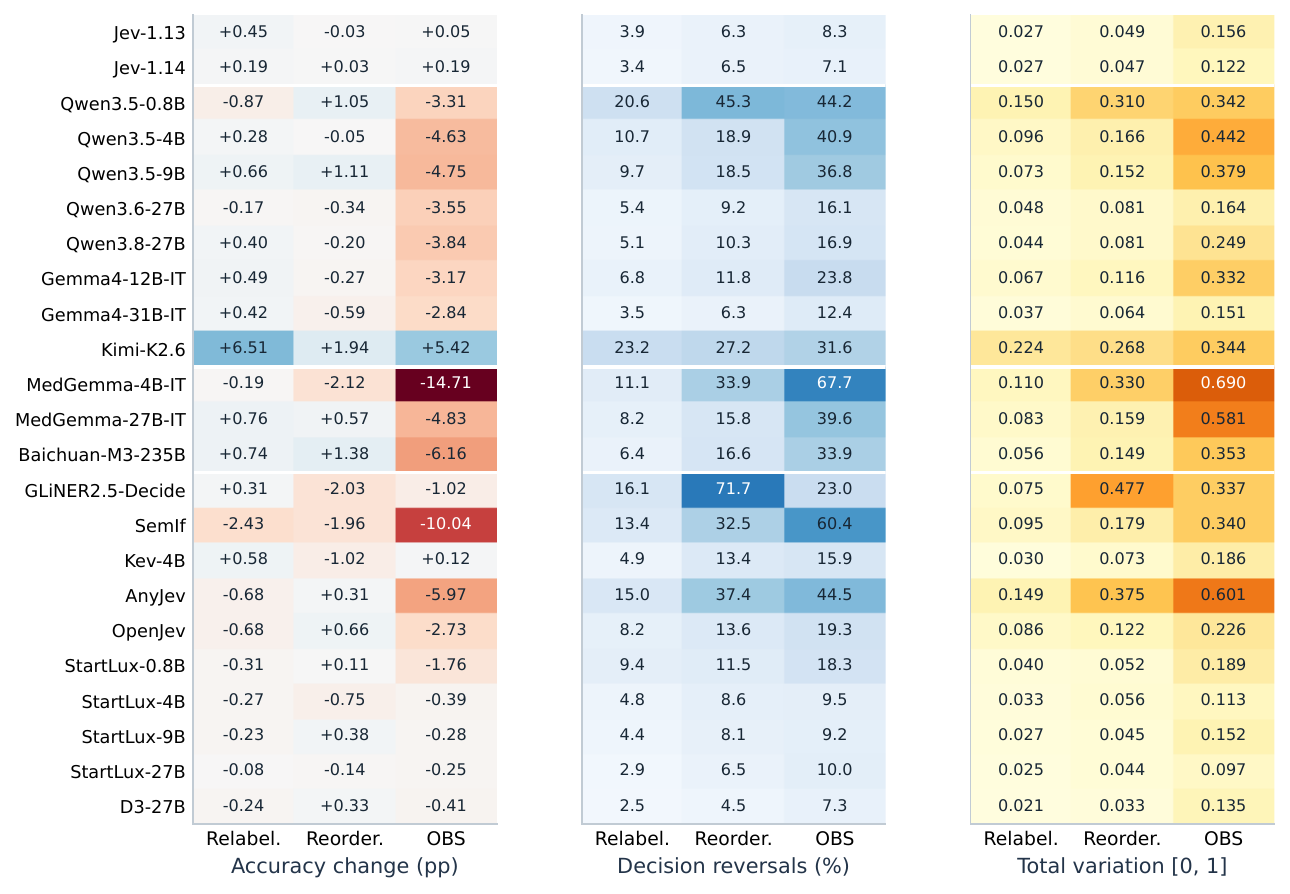}
\caption{Three distinct views of each interface change for all models, with equal dataset weighting. Left: accuracy change relative to original choice, in percentage points. Middle: the percentage of predictions that change after semantic alignment. Right: total-variation distance between aligned candidate distributions. A small accuracy difference can coexist with many flips or a large probability shift.}
\label{fig:paired_drift}
\end{figure}

This section supplements Section~\ref{sec:robustness}. Table~\ref{tab:format_all} reports accuracy, macro-F1, and prediction reversal after semantic alignment; Figure~\ref{fig:paired_drift} adds probability changes on the same cohorts. All averages weight datasets equally. Relabeling and reordering preserve option meanings, whereas OBS changes probability construction. 
The resulting probability distance captures differences between the joint-choice and option-wise scoring formulations.
Accuracy-change variance and mean reversal follow the main-text definitions.

\paragraph{Probability-distribution distance.}
For entry $i$ in dataset $d$ and transformation $t$, let $p_{di0}$ denote
the original Choice distribution and let $p_{dit}$ denote the transformed
candidate distribution. Before comparison, $p_{dit}$ is mapped back to the original
candidate identities by the semantic-alignment mapping $\pi_t$. For OBS,
$p_{dit}$ is the normalized distribution obtained from the affirmative
probabilities of the option-wise binary judgments. We compute total
variation distance as
\begin{equation}
\operatorname{TV}_{dit}
=\frac{1}{2}\sum_{k=1}^{K_{di}}
\left|p_{di0,k}-[\pi_t(p_{dit})]_k\right|,
\label{eq:appendix_tv_item}
\end{equation}
where $K_{di}$ is the number of candidates for entry $i$. A value of zero
indicates identical aligned distributions, whereas one indicates disjoint
probability mass. To match the other robustness summaries, we first average
over the $N_d$ entries in each dataset and then average the $D=15$ dataset
means equally:
\begin{equation}
\operatorname{TV}_{t}
=\frac{1}{D}\sum_{d=1}^{D}
\frac{1}{N_d}\sum_{i=1}^{N_d}\operatorname{TV}_{dit}.
\label{eq:appendix_tv_aggregate}
\end{equation}
Thus, the reported distance measures how strongly the candidate-probability
distribution changes under each evaluated interface transformation after
candidate identities have been aligned.

\end{multicols}
\begin{table}[H]
\centering\footnotesize\setlength{\tabcolsep}{3pt}
\begin{tabular*}{\textwidth}{@{\extracolsep{\fill}}lrrrrrrr@{}}
\toprule
Model & Original & Relabeling & Reordering & OBS & $F_{\mathrm{Rel}}$ & $F_{\mathrm{Reo}}$& $F_{\mathrm{OBS}}$ \\
\midrule
\multicolumn{8}{@{}l}{\textit{Jev}} \\
Jev-1.13 & 61.6 / 58.7 & 62.1 / 58.9 & 61.6 / 58.4 & 61.7 / 58.5 & 3.9 & 6.3 & 8.3 \\
Jev-1.14 & 62.3 / 59.7 & 62.5 / 60.0 & 62.3 / 60.0 & 62.5 / 60.1 & 3.4 & 6.5 & 7.1 \\
\midrule
\multicolumn{8}{@{}l}{\textit{General-purpose LLMs}} \\
Qwen3.5-0.8B & 37.6 / 27.2 & 36.8 / 26.6 & 38.7 / 31.9 & 34.3 / 19.0 & 20.6 & 45.3 & 44.2 \\
Qwen3.5-4B & 53.2 / 47.8 & 53.5 / 48.2 & 53.2 / 48.3 & 48.6 / 38.8 & 10.7 & 18.9 & 40.9 \\
Qwen3.5-9B & 55.5 / 52.1 & 56.2 / 52.7 & 56.6 / 53.3 & 50.8 / 43.6 & 9.7 & 18.5 & 36.8 \\
Qwen3.6-27B & 64.0 / 61.2 & 63.8 / 61.2 & 63.6 / 60.6 & 60.4 / 56.6 & 5.4 & 9.2 & 16.1 \\
Qwen3.8-27B & 63.8 / 61.3 & 64.2 / 61.6 & 63.6 / 60.9 & 60.0 / 55.1 & 5.1 & 10.3 & 16.9 \\
Gemma4-12B-IT & 57.9 / 54.4 & 58.4 / 54.9 & 57.7 / 54.2 & 54.8 / 50.0 & 6.8 & 11.8 & 23.8 \\
Gemma4-31B-IT & 66.3 / 62.1 & 66.8 / 62.4 & 65.8 / 61.7 & 63.5 / 59.5 & 3.5 & 6.3 & 12.4 \\
Kimi-K2.6 & 54.2 / 50.0 & 60.7 / 57.4 & 56.1 / 53.3 & 59.6 / 56.3 & 23.2 & 27.2 & 31.6 \\
\midrule
\multicolumn{8}{@{}l}{\textit{Medical LLMs}} \\
MedGemma-4B-IT & 48.7 / 41.4 & 48.5 / 41.5 & 46.6 / 42.8 & 34.0 / 21.2 & 11.1 & 33.9 & 67.7 \\
MedGemma-27B-IT & 53.7 / 48.5 & 54.4 / 49.8 & 54.2 / 49.1 & 48.8 / 37.8 & 8.2 & 15.8 & 39.6 \\
Baichuan-M3-235B & 56.3 / 53.0 & 57.0 / 53.3 & 57.6 / 53.8 & 50.1 / 43.1 & 6.4 & 16.6 & 33.9 \\
\midrule
\multicolumn{8}{@{}l}{\textit{Jev-like models}} \\
GLiNER2.5-Decide & 32.5 / 17.6 & 32.8 / 19.0 & 30.4 / 24.6 & 31.4 / 15.3 & 16.1 & 71.7 & 23.0 \\
SemIf & 51.5 / 47.7 & 49.1 / 44.0 & 49.6 / 45.3 & 41.5 / 31.5 & 13.4 & 32.5 & 60.4 \\
Kev-4B & 55.3 / 52.4 & 55.9 / 53.1 & 54.3 / 51.1 & 55.4 / 51.9 & 4.9 & 13.4 & 15.9 \\
AnyJev & 47.8 / 42.8 & 47.1 / 41.9 & 48.1 / 43.8 & 41.8 / 33.1 & 15.0 & 37.4 & 44.5 \\
OpenJev & 61.3 / 57.3 & 60.7 / 56.5 & 62.0 / 58.2 & 58.6 / 54.4 & 8.2 & 13.6 & 19.3 \\
StartLux-0.8B & 45.9 / 38.6 & 45.5 / 38.3 & 46.0 / 38.8 & 44.1 / 36.5 & 9.4 & 11.5 & 18.3 \\
StartLux-4B & 58.6 / 55.4 & 58.3 / 55.1 & 57.9 / 54.6 & 58.2 / 54.9 & 4.8 & 8.6 & 9.5 \\
StartLux-9B & 60.0 / 57.2 & 59.8 / 57.0 & 60.4 / 57.7 & 59.7 / 56.7 & 4.4 & 8.1 & 9.2 \\
StartLux-27B & 64.8 / 61.1 & 64.8 / 61.0 & 64.7 / 61.0 & 64.6 / 61.3 & 2.9 & 6.5 & 10.0 \\
D3-27B & 66.5 / 62.3 & 66.3 / 62.2 & 66.9 / 62.7 & 66.1 / 62.0 & 2.5 & 4.5 & 7.3 \\
\bottomrule
\end{tabular*}

\caption{Complete original, relabeling, reordering, and OBS accuracy / macro-F1 (\%), plus the three paired reversal rates, for all 23 models. All scores use equal dataset weights. The main text adds the within-dataset four-format variance and the three-format mean reversal.}
\label{tab:format_all}
\end{table}

\clearpage
\begin{multicols}{2}

\section{Abstention Awareness: Clinical Judgment Details}
\label{app:abstention}

\end{multicols}
\begin{table}[H]
\centering
\small
\setlength{\tabcolsep}{5pt}

\begin{tabular*}{\textwidth}{@{\extracolsep{\fill}}lrrrrrr@{}}
\toprule
Model & Original Acc & NONE Acc & $\Delta$ Acc & Slot flip & Correctness flip & \shortstack[r]{NONE Acc\\(orig. correct)} \\
\midrule
\multicolumn{7}{@{}l}{\textit{Jev}} \\
Jev-1.13 & 77.6 & 65.6 & -12.0 & 18.4 & 16.2 & 70.1 \\
Jev-1.14 & 77.7 & 66.3 & -11.4 & 18.2 & 16.0 & 71.0 \\
\midrule
\multicolumn{7}{@{}l}{\textit{General-purpose LLMs}} \\
Qwen3.5-0.8B & 46.5 & 34.7 & -11.8 & 52.3 & 47.9 & 48.5 \\
Qwen3.5-4B & 66.5 & 51.9 & -14.5 & 29.8 & 26.2 & 51.9 \\
Qwen3.5-9B & 69.6 & 59.3 & -10.4 & 24.4 & 21.1 & 62.1 \\
Qwen3.6-27B & 78.9 & 69.0 & -9.9 & 16.2 & 15.4 & 73.7 \\
Qwen3.8-27B & 78.6 & 68.4 & -10.2 & 17.5 & 15.8 & 72.8 \\
Gemma4-12B-IT & 73.1 & 64.6 & -8.5 & 21.4 & 18.6 & 70.7 \\
Gemma4-31B-IT & 82.3 & 65.9 & -16.4 & 22.6 & 22.0 & 70.5 \\
Kimi-K2.6 & 75.1 & 62.9 & -12.2 & 20.2 & 18.0 & 69.2 \\
\midrule
\multicolumn{7}{@{}l}{\textit{Medical LLMs}} \\
MedGemma-4B-IT & 59.3 & 4.7 & -54.6 & 60.6 & 54.8 & 5.9 \\
MedGemma-27B-IT & 69.6 & 46.6 & -23.0 & 30.1 & 27.9 & 49.4 \\
Baichuan-M3-235B & 72.1 & 51.1 & -21.0 & 28.1 & 25.8 & 54.8 \\
\midrule
\multicolumn{7}{@{}l}{\textit{Jev-like models}} \\
GLiNER2.5-Decide & 24.6 & 40.4 & +15.8 & 26.5 & 22.4 & 87.5 \\
SemIf & 61.9 & 35.9 & -26.0 & 36.4 & 33.9 & 42.1 \\
Kev-4B & 67.0 & 77.4 & +10.4 & 21.4 & 21.1 & 86.7 \\
AnyJev & 58.4 & 40.2 & -18.2 & 32.5 & 27.8 & 47.6 \\
OpenJev & 75.8 & 64.7 & -11.1 & 20.6 & 18.2 & 70.2 \\
StartLux-0.8B & 54.6 & 33.1 & -21.5 & 31.4 & 29.1 & 38.6 \\
StartLux-4B & 71.3 & 60.0 & -11.2 & 20.1 & 16.9 & 64.2 \\
StartLux-9B & 75.7 & 60.9 & -14.8 & 20.8 & 19.6 & 64.5 \\
StartLux-27B & 83.3 & 66.8 & -16.5 & 22.4 & 21.2 & 69.5 \\
D3-27B & 82.5 & 67.0 & -15.5 & 20.7 & 19.3 & 70.1 \\
\bottomrule
\end{tabular*}

\caption{Correct-answer replacement across four Clinical Judgment
datasets (2,382 paired entries).
Accuracy and reversal rates are percentages. $\Delta$ Acc is
NONE accuracy minus original accuracy in percentage points.
NONE Acc (orig. correct) measures accuracy after replacement
on items each model originally answered correctly.}
\label{tab:none_full}
\end{table}

\begin{table}[H]
\centering
\small
\setlength{\tabcolsep}{4pt}
\begin{tabular*}{\textwidth}{@{\extracolsep{\fill}}lrrrrrr@{}}
\toprule

& \multicolumn{2}{c}{Original Choice}
& \multicolumn{2}{c}{With Abstention}
& \multicolumn{2}{c}{Gains} \\
\cmidrule(lr){2-3}
\cmidrule(lr){4-5}
\cmidrule(l){6-7}
Model
& $A^0_{\mathrm{all}}$
& $A^0_S$
& $R$
& $A^+_S$
& $\Delta$
& $\Delta_S$ \\
\midrule
\multicolumn{7}{@{}l}{\textit{Jev}} \\
Jev-1.13 & 74.9 & 75.5 & 4.5 & 75.2 & +0.3 & -0.3 \\
Jev-1.14 & 76.8 & 77.4 & 4.4 & 77.3 & +0.5 & -0.1 \\
\midrule
\multicolumn{7}{@{}l}{\textit{General-purpose LLMs}} \\
Qwen3.5-0.8B & 39.7 & 44.9 & 35.2 & 44.7 & +5.0 & -0.2 \\
Qwen3.5-4B & 60.4 & 60.8 & 2.8 & 62.4 & +2.0 & +1.6 \\
Qwen3.5-9B & 64.4 & 65.4 & 3.2 & 65.0 & +0.6 & -0.4 \\
Qwen3.6-27B & 72.9 & 73.5 & 2.8 & 73.2 & +0.3 & -0.3 \\
Qwen3.8-27B & 72.9 & 73.4 & 2.3 & 74.1 & +1.3 & +0.8 \\
Gemma4-12B-IT & 69.9 & 71.0 & 5.0 & 71.7 & +1.9 & +0.7 \\
Gemma4-31B-IT & 80.1 & 80.9 & 4.2 & 80.7 & +0.6 & -0.1 \\
Kimi-K2.6 & 65.7 & 70.5 & 41.3 & 71.4 & +5.7 & +1.0 \\
\midrule
\multicolumn{7}{@{}l}{\textit{Medical LLMs}} \\
MedGemma-4B-IT & 56.3 & 56.6 & 0.8 & 56.2 & -0.1 & -0.4 \\
MedGemma-27B-IT & 63.6 & 64.3 & 2.1 & 64.0 & +0.5 & -0.3 \\
Baichuan-M3-235B & 65.9 & 66.3 & 1.7 & 67.6 & +1.8 & +1.3 \\
\midrule
\multicolumn{7}{@{}l}{\textit{Jev-like models}} \\
GLiNER2.5-Decide & 28.7 & 25.4 & 59.9 & 25.3 & -3.5 & -0.2 \\
SemIf & 58.8 & 59.9 & 11.9 & 56.4 & -2.4 & -3.5 \\
Kev-4B & 63.8 & 68.9 & 23.9 & 67.9 & +4.1 & -1.1 \\
AnyJev & 53.8 & 54.3 & 5.3 & 51.9 & -1.9 & -2.4 \\
OpenJev & 71.1 & 71.9 & 4.1 & 72.7 & +1.6 & +0.8 \\
StartLux-0.8B & 49.1 & 53.2 & 33.1 & 55.8 & +6.7 & +2.6 \\
StartLux-4B & 69.2 & 70.2 & 6.0 & 69.8 & +0.7 & -0.3 \\
StartLux-9B & 73.1 & 73.8 & 4.7 & 73.4 & +0.2 & -0.4 \\
StartLux-27B & 78.9 & 79.7 & 5.6 & 78.2 & -0.7 & -1.5 \\
D3-27B & 79.2 & 79.7 & 2.9 & 78.9 & -0.4 & -0.8 \\
\bottomrule
\end{tabular*}

\caption{Optional abstention across six Clinical Judgment datasets
(2,501 entries).
$A_{\mathrm{all}}^0$ and $A_S^0$ are original accuracy on the full
cohort and retained subset. $R$ is the mean abstention rate and
$A_S^+$ the mean retained accuracy (all in \%).
Gains $\Delta$ and $\Delta^S$ are in percentage points, computed
before rounding.}
\label{tab:abstention_B_all}
\end{table}
\begin{multicols}{2}

As in Section~\ref{sec:boundaries}, this analysis uses direct choice only. Correct-answer replacement covers four Clinical Judgment datasets (2,382 entries) and is reported in Table~\ref{tab:none_full}. Table~\ref{tab:abstention} below supplements the optional-abstention comparison in Table~\ref{tab:abstention_B}, using all six Clinical Judgment datasets (2,501 entries). Both experiments average dataset-level rates equally. The optional-abstention full-cohort baseline thus matches the Clinical Judgment score in Table~\ref{tab:overall}. 
Retained counts are pooled across datasets, while accuracy and abstention rates are averaged equally across datasets.

\end{multicols}
\begin{table}[H]
\centering\footnotesize
\setlength{\tabcolsep}{4pt}
\begin{tabular*}{\textwidth}{@{\extracolsep{\fill}}lrrrrr@{}}
\toprule
Model & Abstain & Retained Acc/F1 & Original: all & Original: retained & Retained $N$ \\
\midrule
\multicolumn{6}{@{}l}{\textit{Jev}} \\
Jev-1.13 & 4.5 & 75.2 / 74.5 & 74.9 / 74.3 & 75.5 / 74.9 & 2241 \\
Jev-1.14 & 4.4 & 77.3 / 76.6 & 76.8 / 76.2 & 77.4 / 76.8 & 2270 \\
\midrule
\multicolumn{6}{@{}l}{\textit{General-purpose LLMs}} \\
Qwen3.5-0.8B & 35.2 & 44.7 / 38.6 & 39.7 / 35.0 & 44.9 / 39.9 & 1469 \\
Qwen3.5-4B & 2.8 & 62.4 / 62.0 & 60.4 / 59.7 & 60.8 / 60.1 & 2333 \\
Qwen3.5-9B & 3.2 & 65.0 / 63.3 & 64.4 / 62.7 & 65.4 / 63.8 & 2330 \\
Qwen3.6-27B & 2.8 & 73.2 / 71.6 & 72.9 / 71.5 & 73.5 / 72.2 & 2336 \\
Qwen3.8-27B & 2.3 & 74.1 / 73.6 & 72.9 / 72.5 & 73.4 / 73.0 & 2363 \\
Gemma4-12B-IT & 5.0 & 71.7 / 70.8 & 69.9 / 69.0 & 71.0 / 70.1 & 2207 \\
Gemma4-31B-IT & 4.2 & 80.7 / 80.6 & 80.1 / 79.8 & 80.9 / 80.5 & 2261 \\
Kimi-K2.6 & 41.3 & 71.4 / 62.4 & 65.7 / 61.0 & 70.5 / 61.6 & 774 \\
\midrule
\multicolumn{6}{@{}l}{\textit{Medical LLMs}} \\
MedGemma-4B-IT & 0.8 & 56.2 / 52.2 & 56.3 / 52.7 & 56.6 / 53.0 & 2461 \\
MedGemma-27B-IT & 2.1 & 64.0 / 62.2 & 63.6 / 61.7 & 64.3 / 62.5 & 2376 \\
Baichuan-M3-235B & 1.7 & 67.6 / 65.6 & 65.9 / 64.1 & 66.3 / 64.5 & 2406 \\
\midrule
\multicolumn{6}{@{}l}{\textit{Jev-like models}} \\
GLiNER2.5-Decide & 59.9 & 25.3 / 15.3 & 28.7 / 17.5 & 25.4 / 15.7 & 633 \\
SemIf & 11.9 & 56.4 / 53.5 & 58.8 / 57.4 & 59.9 / 57.8 & 2162 \\
Kev-4B & 23.9 & 67.9 / 67.0 & 63.8 / 63.2 & 68.9 / 68.2 & 1846 \\
AnyJev & 5.3 & 51.9 / 47.8 & 53.8 / 51.9 & 54.3 / 51.7 & 2350 \\
OpenJev & 4.1 & 72.7 / 71.5 & 71.1 / 70.0 & 71.9 / 70.8 & 2255 \\
StartLux-0.8B & 33.1 & 55.8 / 54.2 & 49.1 / 47.3 & 53.2 / 51.4 & 1216 \\
StartLux-4B & 6.0 & 69.8 / 69.3 & 69.2 / 68.9 & 70.2 / 69.8 & 2151 \\
StartLux-9B & 4.7 & 73.4 / 72.5 & 73.1 / 72.4 & 73.8 / 73.1 & 2225 \\
StartLux-27B & 5.6 & 78.2 / 77.6 & 78.9 / 78.1 & 79.7 / 78.9 & 2178 \\
D3-27B & 2.9 & 78.9 / 78.5 & 79.2 / 78.9 & 79.7 / 79.4 & 2327 \\
\bottomrule
\end{tabular*}

\caption{All-model \textbf{direct-choice} abstention comparison in B only (six datasets; 2,501 entries). Accuracy, macro-F1, and abstention rates are computed within each dataset and averaged equally. Original: all uses choice without an abstention option on the complete B cohort; Original: retained uses that protocol on exactly the modified version's retained items. Retained $N$ is the pooled count across datasets and excludes refusals. Retained-set metrics omit a dataset only when no entries are retained there (GLiNER2.5-Decide on SCT).}
\label{tab:abstention}
\end{table}
\begin{multicols}{2}

Comparing retained accuracy with ``Original: retained'' gives $\Delta^S$, the change on the same retained items; comparing it with ``Original: all'' gives $\Delta$, the change relative to the full-cohort baseline. For Jev-1.13, these differences are $-0.29$ and $+0.33$ percentage points, respectively: the small full-cohort gain accompanies selection of easier subsets, while decisions on those subsets become slightly less accurate. For GLiNER2.5-Decide, $\Delta$ additionally reflects the exclusion of SCT from the retained-accuracy mean; $\Delta^S$ still compares the same five datasets. 

\end{multicols}
\clearpage
\begin{multicols}{2}

\section{Probability-Based Reliability: Full Threshold Comparison}
\label{app:confidence}
\paragraph{Selection metrics.}
On the pooled cohort of $N=8{,}742$ entries, $c_i$ is the probability assigned to the saved Choice decision or the largest normalized OBS probability. With accepted reference set $G_i$, threshold selection uses
\begin{align}
S_\tau&=\{i:c_i\geq\tau\}, \qquad C_\tau=|S_\tau|/N,\notag\\
A_\tau&=\frac{1}{|S_\tau|}\sum_{i\in S_\tau}\mathbf{1}[\hat y_i\in G_i].
\label{eq:confidence_selection}
\end{align}
For fixed global coverage, the same metrics apply to the highest-confidence $\lfloor Nr\rfloor$ entries, where $r\in\{0.05,0.10,0.15,0.20,0.25\}$. Coverage fixes the retained count, not the selected items or their task composition.

\end{multicols}
\begin{figure}[H]
\centering
\includegraphics[width=\textwidth]{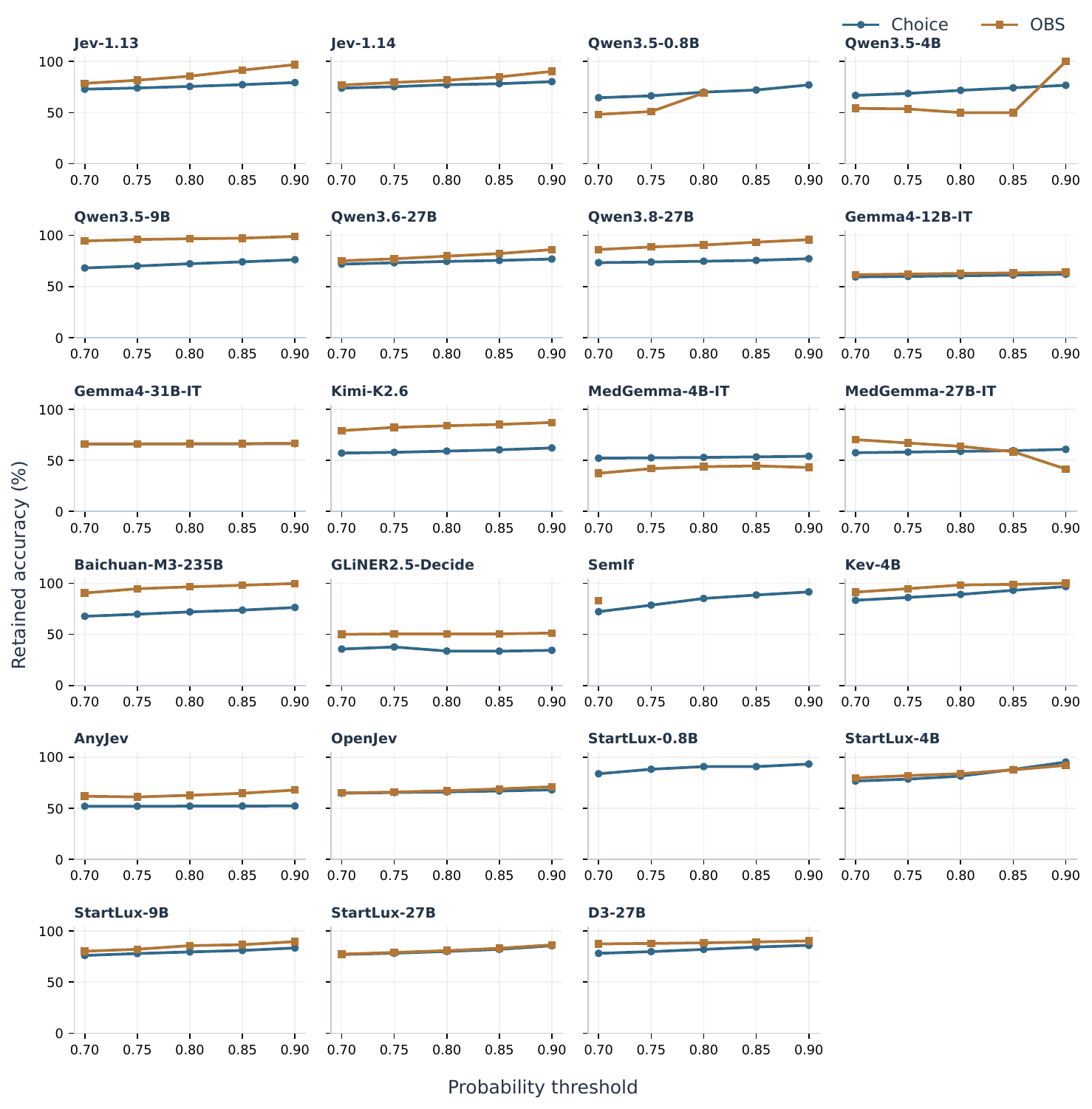}
\caption{Retained accuracy across confidence thresholds for all
23 models. Blue circles denote Choice; orange squares denote
normalized OBS. Missing points indicate empty retained sets.
Coverage and counts are reported in Figure~\ref{fig:coverage}
and Table~\ref{tab:threshold_counts}.}
\label{fig:thresholds}
\end{figure}

\begin{table}[H]
\centering
\small
\setlength{\tabcolsep}{4.5pt}
\begin{tabular}{lrrrr}
\toprule
Model & Choice Acc & Cov. & OBS Acc & Cov. \\
\midrule
\multicolumn{5}{@{}l}{\textit{Jev}} \\
Jev-1.13 & 79.4 & 42.3 & 97.0 & 18.8 \\
Jev-1.14 & 80.3 & 39.9 & 90.3 & 24.7 \\
\midrule
\multicolumn{5}{@{}l}{\textit{General-purpose LLMs}} \\
Qwen3.5-0.8B & 77.1 & 3.1 & -- & 0.0 \\
Qwen3.5-4B & 76.7 & 44.0 & 100.0 & 0.0 \\
Qwen3.5-9B & 76.2 & 42.5 & 99.0 & 2.3 \\
Qwen3.6-27B & 76.9 & 50.1 & 86.1 & 33.1 \\
Qwen3.8-27B & 77.2 & 45.9 & 95.9 & 17.3 \\
Gemma4-12B-IT & 62.0 & 87.2 & 64.0 & 57.7 \\
Gemma4-31B-IT & 66.5 & 96.9 & 66.6 & 83.9 \\
Kimi-K2.6 & 62.2 & 66.6 & 87.1 & 33.2 \\
\midrule
\multicolumn{5}{@{}l}{\textit{Medical LLMs}} \\
MedGemma-4B-IT & 54.1 & 84.4 & 43.1 & 14.9 \\
MedGemma-27B-IT & 60.8 & 81.7 & 41.6 & 1.0 \\
Baichuan-M3-235B & 76.3 & 45.6 & 99.8 & 10.4 \\
\midrule
\multicolumn{5}{@{}l}{\textit{Jev-like models}} \\
GLiNER2.5-Decide & 34.3 & 28.2 & 51.2 & 4.2 \\
SemIf & 91.5 & 18.5 & -- & 0.0 \\
Kev-4B & 96.8 & 18.4 & 100.0 & 0.8 \\
AnyJev & 52.2 & 88.3 & 67.8 & 5.3 \\
OpenJev & 68.0 & 73.6 & 71.0 & 48.0 \\
StartLux-0.8B & 93.2 & 0.8 & -- & 0.0 \\
StartLux-4B & 95.2 & 21.4 & 92.0 & 15.6 \\
StartLux-9B & 83.4 & 26.6 & 89.7 & 7.3 \\
StartLux-27B & 85.7 & 32.4 & 86.5 & 25.1 \\
D3-27B & 86.0 & 35.7 & 90.4 & 24.1 \\
\bottomrule
\end{tabular}

\caption{Retained accuracy and coverage (\%) at $\tau=0.90$.
Accuracy pools retained entries; coverage uses all 8,742 entries.
Results for all five thresholds are in Table~\ref{tab:thresholds}.}
\label{tab:confidence090}
\end{table}
\begin{multicols}{2}

This section supplements Section~\ref{sec:confidence} with all five confidence thresholds: 0.70, 0.75, 0.80, 0.85, and 0.90. For each of the 23 model configurations, adjacent rows compare direct choice (C) with normalized option-wise binary scoring (O). Confidence is the probability assigned to the selected candidate, and both protocols use their original candidate sets.

\begin{figure}[H]
\centering
\includegraphics[width=\textwidth]{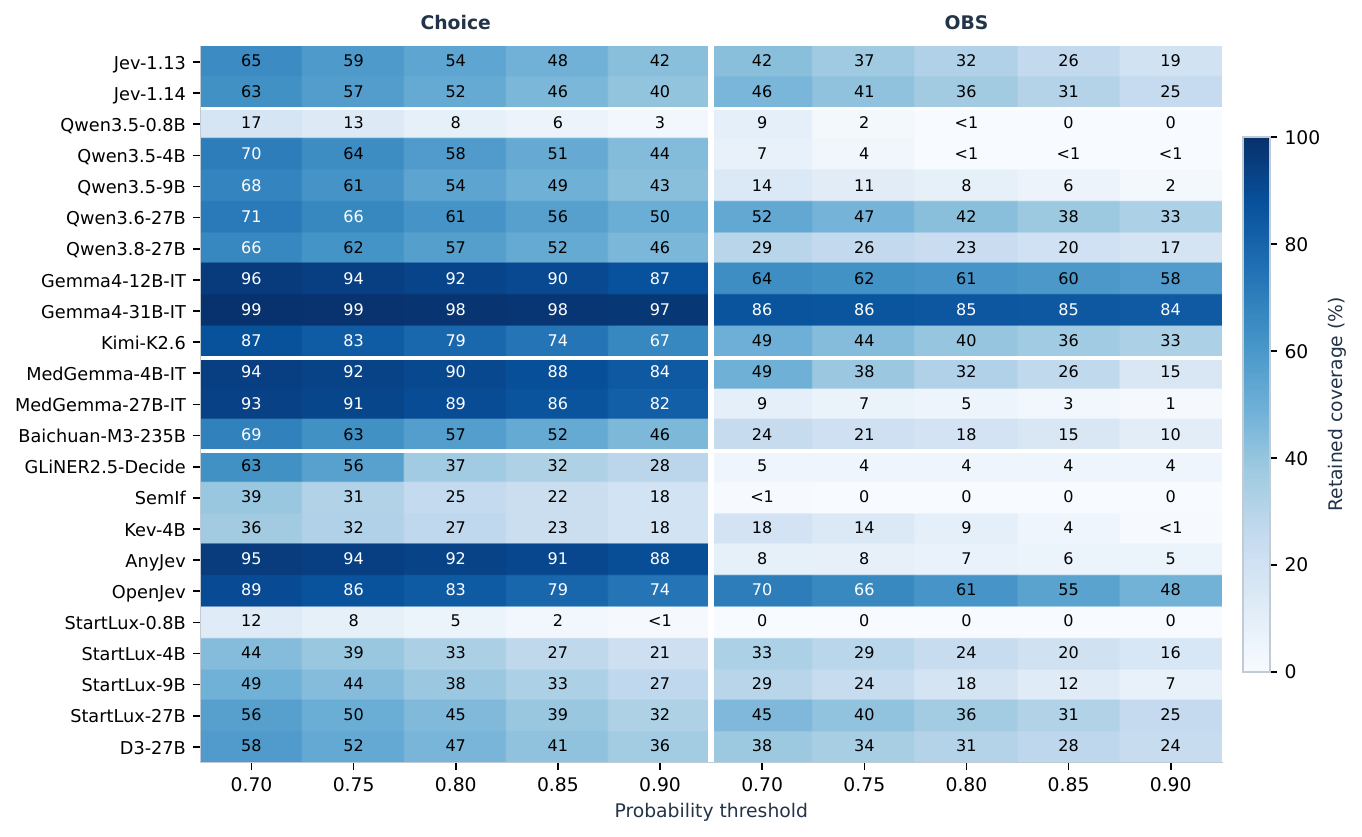}
\caption{Rounded coverage percentages at every threshold for all models; $<1$ denotes nonzero coverage below 1\%. The denominator is 8,742 complete parent items. Accuracy and coverage jointly characterize the operating points of probability-based selection.}
\label{fig:coverage}
\end{figure}

Figure~\ref{fig:thresholds} and Table~\ref{tab:thresholds} report retained accuracy, while Table~\ref{tab:confidence090} summarizes accuracy and coverage at 0.90, and Table~\ref{tab:threshold_counts} reports retained sample counts. Accuracy is pooled across retained entries from all 15 datasets, with undefined accuracy for empty retained subsets indicated by ``--''. Coverage is the retained count divided by 8,742. Comparisons use common confidence thresholds, with coverage and retained subsets determined separately for each model and scoring protocol. Figure~\ref{fig:coverage} illustrates the resulting coverage patterns.

\end{multicols}
\begin{table}[H]
\centering\footnotesize
\setlength{\tabcolsep}{5pt}
\begin{tabular*}{\textwidth}{@{\extracolsep{\fill}}lrrrrrr@{}}
\toprule
Model & Mode & 0.70 & 0.75 & 0.80 & 0.85 & 0.90 \\
\midrule
\multicolumn{7}{@{}l}{\textit{Jev}} \\
Jev-1.13 & C & 72.9 & 74.1 & 75.6 & 77.3 & 79.4 \\
 & OBS & 78.6 & 81.7 & 85.7 & 91.6 & 97.0 \\
Jev-1.14 & C & 74.0 & 75.4 & 77.3 & 78.3 & 80.3 \\
 & OBS & 77.0 & 79.5 & 81.7 & 84.9 & 90.3 \\
\midrule
\multicolumn{7}{@{}l}{\textit{General-purpose LLMs}} \\
Qwen3.5-0.8B & C & 64.6 & 66.4 & 70.0 & 72.1 & 77.1 \\
 & OBS & 48.3 & 51.0 & 69.2 & -- & -- \\
Qwen3.5-4B & C & 66.9 & 68.8 & 71.8 & 74.3 & 76.7 \\
 & OBS & 54.1 & 53.6 & 50.0 & 50.0 & 100.0 \\
Qwen3.5-9B & C & 68.2 & 70.1 & 72.3 & 74.1 & 76.2 \\
 & OBS & 94.6 & 96.0 & 96.8 & 97.3 & 99.0 \\
Qwen3.6-27B & C & 71.9 & 73.2 & 74.6 & 75.5 & 76.9 \\
 & OBS & 75.2 & 77.1 & 79.7 & 82.2 & 86.1 \\
Qwen3.8-27B & C & 73.3 & 74.0 & 74.8 & 75.6 & 77.2 \\
 & OBS & 86.2 & 88.7 & 90.6 & 93.4 & 95.9 \\
Gemma4-12B-IT & C & 59.4 & 59.9 & 60.5 & 61.2 & 62.0 \\
 & OBS & 61.6 & 62.2 & 62.8 & 63.3 & 64.0 \\
Gemma4-31B-IT & C & 66.0 & 66.1 & 66.2 & 66.3 & 66.5 \\
 & OBS & 66.1 & 66.2 & 66.3 & 66.4 & 66.6 \\
Kimi-K2.6 & C & 57.2 & 57.9 & 59.1 & 60.4 & 62.2 \\
 & OBS & 79.1 & 82.3 & 83.9 & 85.2 & 87.1 \\
\midrule
\multicolumn{7}{@{}l}{\textit{Medical LLMs}} \\
MedGemma-4B-IT & C & 52.2 & 52.6 & 52.9 & 53.5 & 54.1 \\
 & OBS & 37.4 & 42.0 & 43.9 & 44.6 & 43.1 \\
MedGemma-27B-IT & C & 57.6 & 58.1 & 58.9 & 59.5 & 60.8 \\
 & OBS & 70.4 & 67.0 & 63.8 & 58.5 & 41.6 \\
Baichuan-M3-235B & C & 67.7 & 69.7 & 71.9 & 73.7 & 76.3 \\
 & OBS & 90.4 & 94.6 & 96.5 & 98.0 & 99.8 \\
\midrule
\multicolumn{7}{@{}l}{\textit{Jev-like models}} \\
GLiNER2.5-Decide & C & 35.6 & 37.6 & 33.6 & 33.5 & 34.3 \\
 & OBS & 50.0 & 50.4 & 50.4 & 50.4 & 51.2 \\
SemIf & C & 72.2 & 78.6 & 85.1 & 88.5 & 91.5 \\
 & OBS & 82.9 & -- & -- & -- & -- \\
Kev-4B & C & 83.3 & 86.1 & 89.1 & 93.1 & 96.8 \\
 & OBS & 91.3 & 94.7 & 98.3 & 98.9 & 100.0 \\
AnyJev & C & 51.9 & 52.0 & 52.0 & 52.2 & 52.2 \\
 & OBS & 61.7 & 61.0 & 62.6 & 64.6 & 67.8 \\
OpenJev & C & 64.7 & 65.4 & 66.0 & 66.9 & 68.0 \\
 & OBS & 64.9 & 65.8 & 67.1 & 68.9 & 71.0 \\
StartLux-0.8B & C & 83.7 & 88.2 & 90.8 & 90.7 & 93.2 \\
 & OBS & -- & -- & -- & -- & -- \\
StartLux-4B & C & 76.6 & 78.5 & 81.4 & 88.0 & 95.2 \\
 & OBS & 79.6 & 81.9 & 83.8 & 87.6 & 92.0 \\
StartLux-9B & C & 76.1 & 78.0 & 79.6 & 81.0 & 83.4 \\
 & OBS & 80.3 & 82.1 & 85.7 & 86.7 & 89.7 \\
StartLux-27B & C & 77.0 & 78.3 & 80.1 & 82.1 & 85.7 \\
 & OBS & 77.4 & 79.1 & 80.8 & 83.2 & 86.5 \\
D3-27B & C & 78.2 & 79.9 & 82.0 & 84.4 & 86.0 \\
 & OBS & 87.4 & 87.9 & 88.5 & 89.3 & 90.4 \\
\bottomrule
\end{tabular*}

\caption{Threshold comparison with adjacent choice (C) and normalized OBS (O) rows. Cells are pooled retained accuracy (\%). Empty subsets have undefined scores (--). 
Retained populations are determined separately for each model and scoring protocol.
Table~\ref{tab:threshold_counts} reports exact counts.}
\label{tab:thresholds}
\end{table}
\begin{table}[H]
\centering\footnotesize\setlength{\tabcolsep}{5pt}
\begin{tabular*}{\textwidth}{@{\extracolsep{\fill}}lrrrrrr@{}}
\toprule
Model & Mode & 0.70 & 0.75 & 0.80 & 0.85 & 0.90 \\
\midrule
\multicolumn{7}{@{}l}{\textit{Jev}} \\
Jev-1.13 & C & 5659 & 5139 & 4683 & 4209 & 3700 \\
 & OBS & 3712 & 3236 & 2782 & 2262 & 1647 \\
Jev-1.14 & C & 5483 & 5019 & 4532 & 4024 & 3484 \\
 & OBS & 4030 & 3576 & 3144 & 2682 & 2163 \\
\midrule
\multicolumn{7}{@{}l}{\textit{General-purpose LLMs}} \\
Qwen3.5-0.8B & C & 1499 & 1111 & 734 & 506 & 271 \\
 & OBS & 793 & 147 & 13 & 0 & 0 \\
Qwen3.5-4B & C & 6122 & 5631 & 5067 & 4486 & 3850 \\
 & OBS & 615 & 306 & 14 & 2 & 1 \\
Qwen3.5-9B & C & 5905 & 5353 & 4741 & 4278 & 3717 \\
 & OBS & 1226 & 980 & 741 & 481 & 198 \\
Qwen3.6-27B & C & 6224 & 5805 & 5336 & 4889 & 4384 \\
 & OBS & 4578 & 4135 & 3691 & 3306 & 2894 \\
Qwen3.8-27B & C & 5812 & 5379 & 4951 & 4530 & 4011 \\
 & OBS & 2544 & 2258 & 2006 & 1760 & 1511 \\
Gemma4-12B-IT & C & 8351 & 8227 & 8043 & 7865 & 7626 \\
 & OBS & 5570 & 5447 & 5325 & 5209 & 5044 \\
Gemma4-31B-IT & C & 8648 & 8616 & 8567 & 8537 & 8471 \\
 & OBS & 7519 & 7479 & 7436 & 7396 & 7336 \\
Kimi-K2.6 & C & 7587 & 7262 & 6874 & 6439 & 5819 \\
 & OBS & 4326 & 3817 & 3481 & 3189 & 2900 \\
\bottomrule
\end{tabular*}

\par\medskip
\begin{tabular*}{\textwidth}{@{\extracolsep{\fill}}lrrrrrr@{}}
\toprule
Model & Mode & 0.70 & 0.75 & 0.80 & 0.85 & 0.90 \\
\midrule
\multicolumn{7}{@{}l}{\textit{Medical LLMs}} \\
MedGemma-4B-IT & C & 8211 & 8075 & 7910 & 7690 & 7375 \\
 & OBS & 4294 & 3343 & 2789 & 2291 & 1306 \\
MedGemma-27B-IT & C & 8144 & 7974 & 7752 & 7514 & 7138 \\
 & OBS & 777 & 579 & 417 & 253 & 89 \\
Baichuan-M3-235B & C & 5996 & 5496 & 4957 & 4521 & 3983 \\
 & OBS & 2120 & 1837 & 1597 & 1324 & 911 \\
\midrule
\multicolumn{7}{@{}l}{\textit{Jev-like models}} \\
GLiNER2.5-Decide & C & 5496 & 4872 & 3206 & 2822 & 2467 \\
 & OBS & 396 & 375 & 375 & 373 & 363 \\
SemIf & C & 3405 & 2746 & 2224 & 1919 & 1616 \\
 & OBS & 41 & 0 & 0 & 0 & 0 \\
Kev-4B & C & 3167 & 2768 & 2340 & 1978 & 1605 \\
 & OBS & 1582 & 1226 & 819 & 373 & 68 \\
AnyJev & C & 8325 & 8211 & 8079 & 7935 & 7721 \\
 & OBS & 724 & 685 & 605 & 540 & 463 \\
OpenJev & C & 7814 & 7560 & 7282 & 6916 & 6432 \\
 & OBS & 6116 & 5744 & 5327 & 4808 & 4199 \\
StartLux-0.8B & C & 1051 & 726 & 455 & 216 & 74 \\
 & OBS & 0 & 0 & 0 & 0 & 0 \\
StartLux-4B & C & 3817 & 3392 & 2911 & 2383 & 1869 \\
 & OBS & 2893 & 2493 & 2082 & 1735 & 1360 \\
StartLux-9B & C & 4261 & 3806 & 3339 & 2901 & 2323 \\
 & OBS & 2538 & 2056 & 1556 & 1034 & 638 \\
StartLux-27B & C & 4896 & 4396 & 3927 & 3398 & 2831 \\
 & OBS & 3960 & 3539 & 3135 & 2731 & 2191 \\
D3-27B & C & 5074 & 4549 & 4086 & 3581 & 3123 \\
 & OBS & 3304 & 2989 & 2691 & 2408 & 2106 \\
\bottomrule
\end{tabular*}

\caption{Exact retained $N$ for the threshold scores in Table~\ref{tab:thresholds}. C: choice; O: normalized OBS. The pre-threshold denominator is 8,742 for every model and protocol.}
\label{tab:threshold_counts}
\end{table}

\clearpage
\clearpage
\begin{multicols}{2}

\section{Fixed-Coverage Confidence Ranking}
\label{app:fixed_coverage}

\end{multicols}
\begin{figure}[H]
\centering
\includegraphics[width=0.72\textwidth]{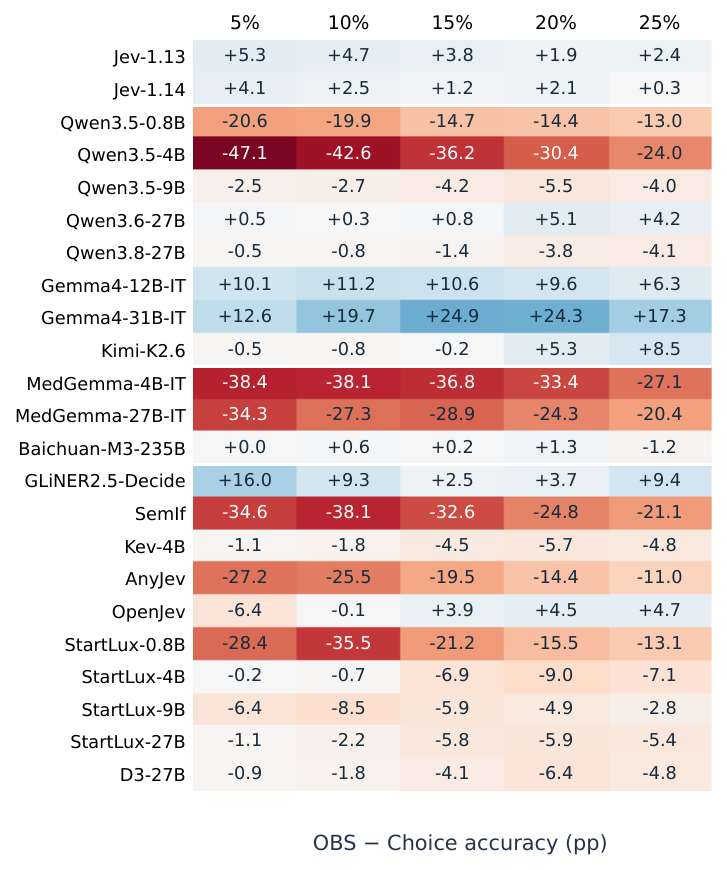}
\caption{OBS minus Choice accuracy (percentage points) at matched
global coverage for all 23 models. Positive values favor OBS.
Absolute accuracies and retained counts are reported in
Table~\ref{tab:fixed_global}.}
\label{fig:fixed_coverage}
\end{figure}
\begin{multicols}{2}

Figure~\ref{fig:fixed_coverage} visualizes OBS-minus-Choice accuracy, and Table~\ref{tab:fixed_global} compares all 22 models at the same retained proportions: the top 5\%, 10\%, 15\%, 20\%, and 25\% of predictions by confidence. These correspond to 437, 874, 1,311, 1,748, and 2,185 entries from the full cohort of 8,742. Unlike fixed thresholds, this comparison holds the number of retained predictions constant across models and protocols.

\paragraph{Jev benefits from OBS at every tested coverage.}
OBS yields higher retained accuracy than Choice at all five coverage levels in both Jev versions. At 5\% coverage, OBS achieves $99.31$--$99.54\%$ accuracy versus $94.05$--$95.42\%$ under Choice; at 25\%, the corresponding ranges are $90.25$--$92.45\%$ and $89.98$--$90.07\%$. The advantage therefore persists beyond the different coverage levels induced by a shared probability threshold, supporting stronger global discrimination under OBS.

\paragraph{The effect varies across models.}

Gemma4-12B-IT, Gemma4-31B-IT, and Qwen3.6-27B benefit from OBS at all five coverage levels. In contrast, D3-27B, SemIf, Kev-4B, AnyJev, and all four StartLux configurations obtain lower retained accuracy under OBS throughout. For D3-27B, Choice versus OBS accuracy is $100.00\%$ versus $99.08\%$ at 5\% coverage and $97.94\%$ versus $91.53\%$ at 20\%. This contrasts with the fixed-threshold comparison, where OBS raises accuracy from $86.04\%$ to $90.36\%$ at $0.90$ while reducing coverage from $35.72\%$ to $24.09\%$. OpenJev also illustrates dependence on the selection level, favoring Choice at 5--10\% coverage but OBS at 15--25\%. Higher accuracy at a shared threshold therefore does not establish better selection at a fixed retention budget.

\end{multicols}
\begin{table*}[p]
\centering\small\setlength{\tabcolsep}{6pt}
\begin{tabular*}{\textwidth}{@{\extracolsep{\fill}}llrrrrr@{}}
\toprule
Model & Mode & 5\% & 10\% & 15\% & 20\% & 25\% \\
\midrule
\multicolumn{7}{@{}l}{\textit{Jev}} \\
Jev-1.13 & C & 94.05 & 93.59 & 93.90 & 94.34 & 90.07 \\
 & OBS & 99.31 & 98.28 & 97.71 & 96.22 & 92.45 \\
Jev-1.14 & C & 95.42 & 96.22 & 96.34 & 93.59 & 89.98 \\
 & OBS & 99.54 & 98.74 & 97.56 & 95.65 & 90.25 \\
\midrule
\multicolumn{7}{@{}l}{\textit{General-purpose LLMs}} \\
Qwen3.5-0.8B & C & 70.48 & 68.65 & 64.99 & 63.39 & 60.73 \\
 & OBS & 49.89 & 48.74 & 50.27 & 48.97 & 47.78 \\
Qwen3.5-4B & C & 99.54 & 99.43 & 98.17 & 94.11 & 88.88 \\
 & OBS & 52.40 & 56.86 & 62.01 & 63.73 & 64.85 \\
Qwen3.5-9B & C & 100.00 & 99.20 & 97.79 & 93.76 & 87.78 \\
 & OBS & 97.48 & 96.45 & 93.59 & 88.27 & 83.80 \\
Qwen3.6-27B & C & 99.54 & 99.31 & 98.47 & 92.96 & 89.79 \\
 & OBS & 100.00 & 99.66 & 99.31 & 98.11 & 93.96 \\
Qwen3.8-27B & C & 100.00 & 99.77 & 98.86 & 97.25 & 93.14 \\
 & OBS & 99.54 & 98.97 & 97.48 & 93.48 & 89.02 \\
Gemma4-12B-IT & C & 88.79 & 87.41 & 87.03 & 84.38 & 82.01 \\
 & OBS & 98.86 & 98.63 & 97.64 & 93.99 & 88.33 \\
Gemma4-31B-IT & C & 86.50 & 78.72 & 72.85 & 71.00 & 74.69 \\
 & OBS & 99.08 & 98.40 & 97.71 & 95.31 & 91.95 \\
Kimi-K2.6 & C & 100.00 & 99.54 & 97.41 & 90.33 & 84.12 \\
 & OBS & 99.54 & 98.74 & 97.25 & 95.65 & 92.59 \\
\midrule
\multicolumn{7}{@{}l}{\textit{Medical LLMs}} \\
MedGemma-4B-IT & C & 82.38 & 82.49 & 79.79 & 76.37 & 72.72 \\
 & OBS & 43.94 & 44.39 & 42.94 & 42.96 & 45.58 \\
MedGemma-27B-IT & C & 99.08 & 97.37 & 95.58 & 92.96 & 87.96 \\
 & OBS & 64.76 & 70.02 & 66.67 & 68.71 & 67.55 \\
Baichuan-M3-235B & C & 100.00 & 99.31 & 97.86 & 93.94 & 90.30 \\
 & OBS & 100.00 & 99.89 & 98.09 & 95.25 & 89.06 \\
\midrule
\multicolumn{7}{@{}l}{\textit{Jev-like models}} \\
GLiNER2.5-Decide & C & 32.27 & 36.27 & 39.21 & 38.04 & 35.24 \\
 & OBS & 48.28 & 45.54 & 41.72 & 41.70 & 44.67 \\
SemIf & C & 99.54 & 97.83 & 95.27 & 90.39 & 85.58 \\
 & OBS & 64.99 & 59.73 & 62.70 & 65.62 & 64.44 \\
Kev-4B & C & 99.77 & 99.66 & 98.93 & 95.54 & 90.66 \\
 & OBS & 98.63 & 97.83 & 94.43 & 89.87 & 85.90 \\
AnyJev & C & 94.51 & 86.38 & 80.17 & 75.23 & 71.17 \\
 & OBS & 67.28 & 60.87 & 60.64 & 60.81 & 60.14 \\
OpenJev & C & 89.24 & 88.56 & 85.96 & 84.78 & 82.84 \\
 & OBS & 82.84 & 88.44 & 89.86 & 89.24 & 87.51 \\
StartLux-0.8B & C & 90.62 & 86.61 & 79.94 & 75.74 & 73.32 \\
 & OBS & 62.24 & 51.14 & 58.73 & 60.24 & 60.18 \\
StartLux-4B & C & 100.00 & 99.89 & 99.39 & 96.40 & 90.48 \\
 & OBS & 99.77 & 99.20 & 92.52 & 87.36 & 83.34 \\
StartLux-9B & C & 99.77 & 96.34 & 92.45 & 88.84 & 84.35 \\
 & OBS & 93.36 & 87.87 & 86.58 & 83.92 & 81.51 \\
StartLux-27B & C & 100.00 & 100.00 & 99.54 & 96.11 & 91.99 \\
 & OBS & 98.86 & 97.83 & 93.75 & 90.16 & 86.54 \\
D3-27B & C & 100.00 & 100.00 & 99.39 & 97.94 & 94.87 \\
 & OBS & 99.08 & 98.17 & 95.27 & 91.53 & 90.07 \\
\bottomrule
\end{tabular*}

\caption{\textbf{Global fixed-coverage selection, all models.} Adjacent rows compare original choice (C) and normalized OBS. Cells are pooled retained accuracy (\%). The five columns retain exactly 437, 874, 1,311, 1,748, and 2,185 entries, respectively, for every row. Task mixtures may differ.}
\label{tab:fixed_global}
\end{table*}

\clearpage
\begin{multicols}{2}

\section{Confidence Distributions and Probability Concentration}
\label{app:distributions}

\end{multicols}
\begin{figure}[H]
\centering\includegraphics[width=\textwidth]{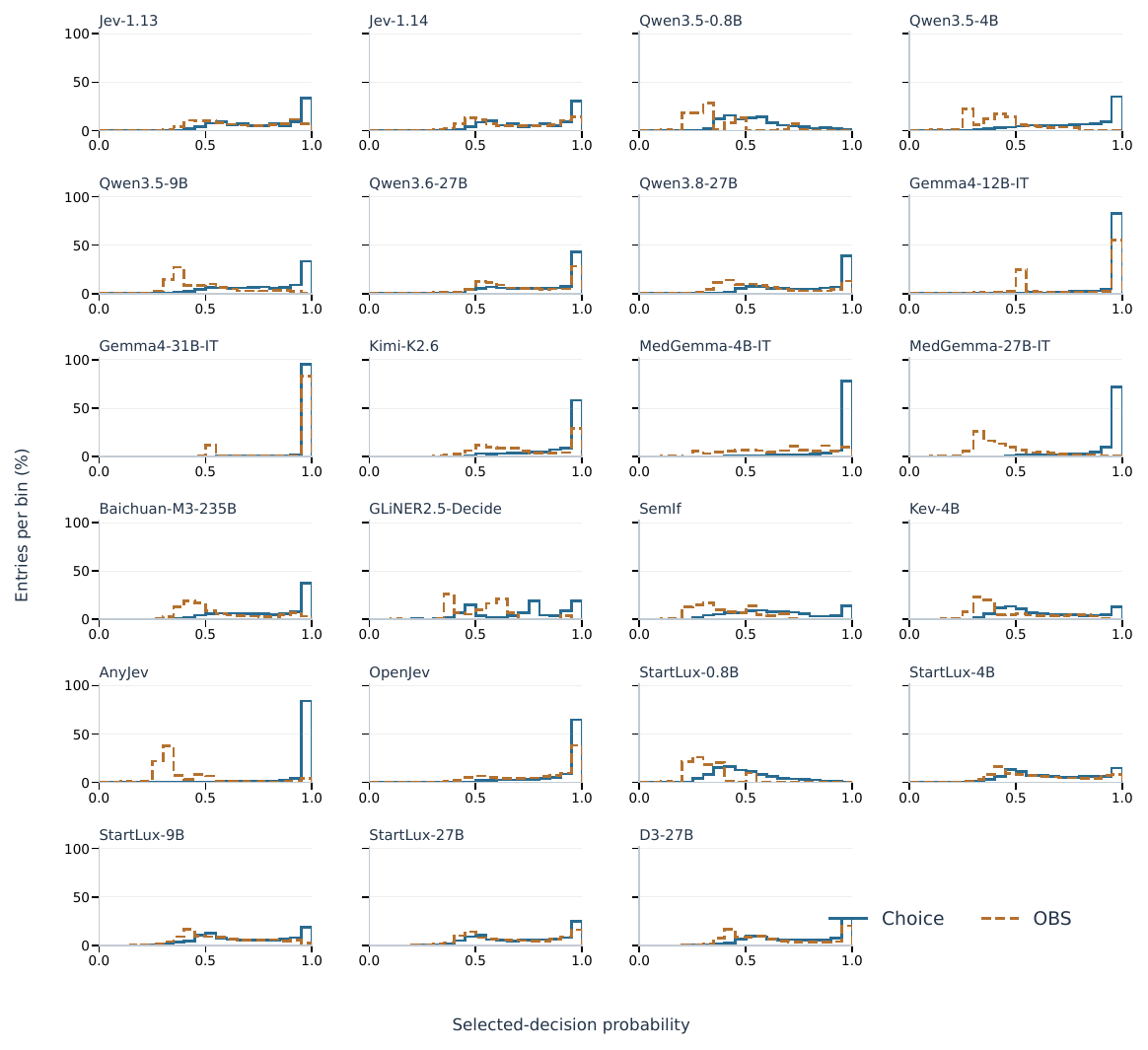}
\caption{\textbf{Choice versus normalized OBS confidence distributions
for all 23 models on the same 8,742 entries.}
Confidence is the probability assigned to the predicted label.
Lines show the percentage of entries in confidence bins of width 0.05.
Blue solid lines denote Choice; orange dashed lines denote OBS.}
\label{fig:confidence_distributions}
\end{figure}
\begin{multicols}{2}

Figure~\ref{fig:confidence_distributions} and Table~\ref{tab:confidence_distribution} compare Choice and OBS confidence distributions across all 23 model configurations on the same 8,742 entries. The summaries distinguish confidence variation across predictions from the concentration of candidate probabilities within each prediction.

\paragraph{OBS lowers confidence while spreading candidate probabilities.}
Across both Jev versions, mean confidence decreases from $0.781$--$0.790$ under Choice to $0.661$--$0.685$ under OBS, while normalized entropy increases from $0.378$--$0.395$ to $0.576$--$0.621$. Thus, OBS assigns less probability to the selected answer and distributes more probability across alternatives. Similar shifts occur in the other evaluated models, but their magnitudes differ: OpenJev's mean confidence decreases from 0.915 to 0.811, whereas AnyJev's decreases from 0.961 to 0.402.

\paragraph{Task mixture matters even without sample selection.}
Across both Jev versions, pooled confidence SD rises from $0.191$--$0.192$ under Choice to $0.198$--$0.202$ under OBS, whereas the equal-dataset mean SD falls from $0.147$--$0.152$ to $0.135$--$0.136$. For Jev-1.13, within-dataset SD declines in 12 of 15 datasets, with ICD-proxy, MedQA, and MIMIC-CDM as the exceptions, while mean normalized entropy increases in every dataset. OBS can therefore make probabilities within an item more uniform while narrowing the spread of confidence within a task. The pooled distribution additionally reflects differences between tasks.

\paragraph{Confidence shape is not reliability.}
Jev-1.13 has 1,819 Choice predictions with confidence exactly one, versus none under OBS. Across both versions, correct-prediction mean confidence falls from $0.837$--$0.847$ under Choice to $0.724$--$0.741$ under OBS, and incorrect-prediction mean confidence falls from $0.690$--$0.700$ to $0.563$--$0.592$. 
Jev's selection advantage is established by its higher retained accuracy at matched coverage, not by lower confidence or higher entropy alone. D3-27B provides a contrasting example. OBS reduces its mean confidence from $0.757$ to $0.654$ and increases mean normalized entropy from $0.430$ to $0.572$, yet lowers retained accuracy at every tested matched coverage level. Less concentrated probabilities therefore do not, by themselves, imply more reliable answer selection.

\end{multicols}
\begin{table*}[p]
\centering\footnotesize\setlength{\tabcolsep}{2pt}
\begin{tabular*}{\textwidth}{@{\extracolsep{\fill}}llrrrrrrr@{}}
\toprule
Model & Mode & Mean $c$ & SD($c$) & IQR($c$) & \shortstack{Mean\\SD$_d$} & \shortstack{Mean\\$H$} & $c_{\rm correct}$ & $c_{\rm wrong}$ \\
\midrule
\multicolumn{9}{@{}l}{\textit{Jev}} \\
Jev-1.13 & C & 0.790 & 0.192 & 0.370 & 0.152 & 0.378 & 0.847 & 0.700 \\
 & OBS & 0.661 & 0.198 & 0.372 & 0.135 & 0.621 & 0.724 & 0.563 \\
Jev-1.14 & C & 0.781 & 0.191 & 0.380 & 0.147 & 0.395 & 0.837 & 0.690 \\
 & OBS & 0.685 & 0.202 & 0.401 & 0.136 & 0.576 & 0.741 & 0.592 \\
\midrule
\multicolumn{9}{@{}l}{\textit{General-purpose LLMs}} \\
Qwen3.5-0.8B & C & 0.553 & 0.151 & 0.200 & 0.092 & 0.785 & 0.605 & 0.515 \\
 & OBS & 0.367 & 0.145 & 0.173 & 0.013 & 0.929 & 0.401 & 0.349 \\
Qwen3.5-4B & C & 0.804 & 0.191 & 0.333 & 0.163 & 0.383 & 0.862 & 0.724 \\
 & OBS & 0.430 & 0.144 & 0.204 & 0.048 & 0.914 & 0.462 & 0.393 \\
Qwen3.5-9B & C & 0.794 & 0.190 & 0.340 & 0.159 & 0.394 & 0.846 & 0.718 \\
 & OBS & 0.487 & 0.168 & 0.204 & 0.075 & 0.806 & 0.550 & 0.411 \\
Qwen3.6-27B & C & 0.827 & 0.184 & 0.340 & 0.150 & 0.324 & 0.861 & 0.760 \\
 & OBS & 0.744 & 0.201 & 0.410 & 0.144 & 0.439 & 0.806 & 0.649 \\
Qwen3.8-27B & C & 0.803 & 0.193 & 0.371 & 0.151 & 0.364 & 0.840 & 0.733 \\
 & OBS & 0.601 & 0.216 & 0.339 & 0.136 & 0.648 & 0.672 & 0.492 \\
Gemma4-12B-IT & C & 0.959 & 0.098 & 0.019 & 0.082 & 0.089 & 0.977 & 0.934 \\
 & OBS & 0.807 & 0.233 & 0.493 & 0.148 & 0.286 & 0.837 & 0.767 \\
Gemma4-31B-IT & C & 0.990 & 0.052 & 0.000 & 0.045 & 0.021 & 0.993 & 0.983 \\
 & OBS & 0.926 & 0.170 & 0.000 & 0.161 & 0.092 & 0.937 & 0.907 \\
Kimi-K2.6 & C & 0.898 & 0.144 & 0.164 & 0.125 & 0.202 & 0.926 & 0.864 \\
 & OBS & 0.736 & 0.204 & 0.421 & 0.132 & 0.435 & 0.800 & 0.632 \\
\midrule
\multicolumn{9}{@{}l}{\textit{Medical LLMs}} \\
MedGemma-4B-IT & C & 0.948 & 0.109 & 0.037 & 0.100 & 0.109 & 0.965 & 0.930 \\
 & OBS & 0.661 & 0.220 & 0.372 & 0.140 & 0.571 & 0.672 & 0.656 \\
MedGemma-27B-IT & C & 0.935 & 0.114 & 0.058 & 0.108 & 0.156 & 0.957 & 0.907 \\
 & OBS & 0.450 & 0.154 & 0.170 & 0.087 & 0.857 & 0.474 & 0.422 \\
Baichuan-M3-235B & C & 0.808 & 0.186 & 0.340 & 0.143 & 0.363 & 0.857 & 0.735 \\
 & OBS & 0.564 & 0.191 & 0.265 & 0.078 & 0.735 & 0.641 & 0.471 \\
\midrule
\multicolumn{9}{@{}l}{\textit{Jev-like models}} \\
GLiNER2.5-Decide & C & 0.724 & 0.206 & 0.432 & 0.070 & 0.545 & 0.752 & 0.711 \\
 & OBS & 0.527 & 0.139 & 0.234 & 0.024 & 0.812 & 0.559 & 0.513 \\
SemIf & C & 0.651 & 0.209 & 0.317 & 0.143 & 0.646 & 0.724 & 0.567 \\
 & OBS & 0.399 & 0.137 & 0.230 & 0.048 & 0.951 & 0.441 & 0.364 \\
Kev-4B & C & 0.639 & 0.207 & 0.355 & 0.133 & 0.638 & 0.723 & 0.527 \\
 & OBS & 0.477 & 0.187 & 0.277 & 0.071 & 0.844 & 0.539 & 0.391 \\
AnyJev & C & 0.961 & 0.099 & 0.009 & 0.100 & 0.080 & 0.967 & 0.956 \\
 & OBS & 0.402 & 0.184 & 0.165 & 0.107 & 0.794 & 0.429 & 0.378 \\
OpenJev & C & 0.915 & 0.133 & 0.109 & 0.121 & 0.176 & 0.933 & 0.883 \\
 & OBS & 0.811 & 0.195 & 0.349 & 0.152 & 0.336 & 0.844 & 0.760 \\
StartLux-0.8B & C & 0.497 & 0.150 & 0.194 & 0.092 & 0.851 & 0.553 & 0.441 \\
 & OBS & 0.322 & 0.086 & 0.099 & 0.018 & 0.992 & 0.340 & 0.304 \\
StartLux-4B & C & 0.679 & 0.201 & 0.368 & 0.132 & 0.581 & 0.750 & 0.578 \\
 & OBS & 0.614 & 0.206 & 0.347 & 0.137 & 0.677 & 0.674 & 0.527 \\
StartLux-9B & C & 0.704 & 0.205 & 0.394 & 0.146 & 0.536 & 0.768 & 0.610 \\
 & OBS & 0.584 & 0.187 & 0.306 & 0.128 & 0.734 & 0.639 & 0.504 \\
StartLux-27B & C & 0.737 & 0.206 & 0.411 & 0.144 & 0.477 & 0.789 & 0.638 \\
 & OBS & 0.682 & 0.208 & 0.414 & 0.151 & 0.590 & 0.729 & 0.596 \\
D3-27B & C & 0.757 & 0.199 & 0.391 & 0.142 & 0.430 & 0.809 & 0.655 \\
 & OBS & 0.654 & 0.223 & 0.436 & 0.144 & 0.572 & 0.719 & 0.523 \\
\bottomrule
\end{tabular*}

\caption{\textbf{All-model distribution summaries on the full paired cohort.} Adjacent rows compare choice (C) and normalized OBS, each with $N=8,742$. All values are on a 0--1 scale. Mean $c$, SD($c$), and IQR($c$) pool entries; Mean SD$_d$ averages within-dataset SDs equally. Mean $H$ is pooled normalized within-item entropy. The last columns give mean confidence conditional on correct or incorrect predictions under that row's protocol; those subsets need not match between rows.}
\label{tab:confidence_distribution}
\end{table*}

\clearpage
\begin{multicols}{2}

\section{Jev Task Coverage under Probability-Based Selection}
\label{app:task_coverage}
Tables~\ref{tab:coverage_family_threshold}--\ref{tab:coverage_dataset_fixed_114} show how selection distributes across task families and datasets. For task $d$, within-task coverage is $|S\cap d|/N_d$, whereas its share of the retained pool is $|S\cap d|/|S|$. The tables report exact retained counts and within-task coverage. Family coverage pools the entries within each family, matching the pooled analysis in Section~\ref{sec:confidence}. All settings use the same selections as the preceding accuracy comparisons. Fixed-coverage selection ranks all 8,742 entries globally, not separately within datasets.

\paragraph{High-probability selections concentrate on Jev's stronger tasks.}
Clinical Judgment has the highest within-family coverage under both scoring interfaces at every tested threshold and global retention level. Across the two Jev versions, at the 0.90 OBS threshold, it retains 1,361--1,613 of 2,501 entries ($54.42$--$64.49\%$) and contributes $74.57$--$82.64\%$ of the retained pool, despite accounting for only 28.61\% of the full cohort. ICD-proxy and MIMIC-CDM together account for $70.46$--$78.08\%$ of that pool, consistent with their high original accuracy. At global top-5\% coverage, 418--425 of the 437 OBS-selected entries ($95.65$--$97.25\%$) belong to Clinical Judgment.

This concentration is not uniform across all text-driven tasks. OBS retains no CTG-21 or MedXpertQA entries at 0.90, and no entries from either dataset at any tested global retention level. Conversely, VitalDB contributes retained entries despite its weaker overall accuracy. Thus, the results indicate concentration in particular task strengths. Global retained-accuracy improvements also reflect changing task composition.

\end{multicols}
\begin{table}[H]
\centering\small
\setlength{\tabcolsep}{4pt}
\begin{tabular}{llrlrrrrr}
\toprule
Family & Task & $N$ & Readout & $0.70$ & $0.75$ & $0.80$ & $0.85$ & $0.90$ \\
\midrule
A & Triage / navigation & 3095 & Choice & 1879 (60.7) & 1607 (51.9) & 1371 (44.3) & 1146 (37.0) & 881 (28.5) \\
 &  &  & OBS & 669 (21.6) & 474 (15.3) & 314 (10.1) & 140 (4.5) & 26 (0.8) \\
\midrule
B & Clinical judgment & 2501 & Choice & 2235 (89.4) & 2183 (87.3) & 2127 (85.0) & 2065 (82.6) & 1997 (79.8) \\
 &  &  & OBS & 1994 (79.7) & 1902 (76.0) & 1811 (72.4) & 1660 (66.4) & 1361 (54.4) \\
\midrule
C & Safety review & 1812 & Choice & 1139 (62.9) & 1006 (55.5) & 873 (48.2) & 713 (39.3) & 557 (30.7) \\
 &  &  & OBS & 742 (40.9) & 580 (32.0) & 422 (23.3) & 274 (15.1) & 132 (7.3) \\
\midrule
D & Physiological assessment & 1334 & Choice & 406 (30.4) & 343 (25.7) & 312 (23.4) & 285 (21.4) & 265 (19.9) \\
 &  &  & OBS & 307 (23.0) & 280 (21.0) & 235 (17.6) & 188 (14.1) & 128 (9.6) \\
\bottomrule
\end{tabular}
\caption{Jev 1.13 coverage by task family at probability thresholds. Cells report retained counts followed by within-task coverage (\%) in parentheses. The denominator is the row's $N$, not the total selected pool. Choice and OBS use their own selections.}
\label{tab:coverage_family_threshold}
\end{table}

\clearpage
\begin{table}[H]
\centering\small
\setlength{\tabcolsep}{4pt}
\begin{tabular}{llrlrrrrr}
\toprule
Family & Task & $N$ & Readout & $0.70$ & $0.75$ & $0.80$ & $0.85$ & $0.90$ \\
\midrule
A & Triage / navigation & 3095 & Choice & 1853 (59.9) & 1634 (52.8) & 1382 (44.7) & 1093 (35.3) & 815 (26.3) \\
 &  &  & OBS & 872 (28.2) & 673 (21.7) & 492 (15.9) & 325 (10.5) & 151 (4.9) \\
\midrule
B & Clinical judgment & 2501 & Choice & 2233 (89.3) & 2174 (86.9) & 2113 (84.5) & 2052 (82.0) & 1969 (78.7) \\
 &  &  & OBS & 2049 (81.9) & 1976 (79.0) & 1886 (75.4) & 1773 (70.9) & 1613 (64.5) \\
\midrule
C & Safety review & 1812 & Choice & 1079 (59.5) & 919 (50.7) & 770 (42.5) & 629 (34.7) & 472 (26.0) \\
 &  &  & OBS & 812 (44.8) & 661 (36.5) & 521 (28.8) & 370 (20.4) & 213 (11.8) \\
\midrule
D & Physiological assessment & 1334 & Choice & 318 (23.8) & 292 (21.9) & 267 (20.0) & 250 (18.7) & 228 (17.1) \\
 &  &  & OBS & 297 (22.3) & 266 (19.9) & 245 (18.4) & 214 (16.0) & 186 (13.9) \\
\bottomrule
\end{tabular}
\caption{Jev 1.14 coverage by task family at probability thresholds. Cells report retained counts followed by within-task coverage (\%) in parentheses. The denominator is the row's $N$, not the total selected pool. Choice and OBS use their own selections.}
\label{tab:coverage_family_threshold_114}
\end{table}

\begin{table}[H]
\centering\small
\setlength{\tabcolsep}{4pt}
\begin{tabular}{llrlrrrrr}
\toprule
Family & Task & $N$ & Readout & 5\% & 10\% & 15\% & 20\% & 25\% \\
\midrule
A & Triage / navigation & 3095 & Choice & 34 (1.1) & 69 (2.2) & 100 (3.2) & 123 (4.0) & 256 (8.3) \\
 &  &  & OBS & 0 (0.0) & 3 (0.1) & 12 (0.4) & 37 (1.2) & 117 (3.8) \\
\midrule
B & Clinical judgment & 2501 & Choice & 355 (14.2) & 699 (27.9) & 1052 (42.1) & 1411 (56.4) & 1614 (64.5) \\
 &  &  & OBS & 418 (16.7) & 823 (32.9) & 1159 (46.3) & 1417 (56.7) & 1635 (65.4) \\
\midrule
C & Safety review & 1812 & Choice & 22 (1.2) & 47 (2.6) & 70 (3.9) & 84 (4.6) & 148 (8.2) \\
 &  &  & OBS & 19 (1.0) & 47 (2.6) & 85 (4.7) & 149 (8.2) & 253 (14.0) \\
\midrule
D & Physiological assessment & 1334 & Choice & 26 (1.9) & 59 (4.4) & 89 (6.7) & 130 (9.7) & 167 (12.5) \\
 &  &  & OBS & 0 (0.0) & 1 (0.1) & 55 (4.1) & 145 (10.9) & 180 (13.5) \\
\bottomrule
\end{tabular}
\caption{Jev 1.13 coverage by task family under global top-percentage selection. Columns denote global retained proportions; they do not impose equal coverage within tasks. Cells report retained counts followed by within-task coverage (\%) in parentheses. The denominator is the row's $N$, not the total selected pool. Choice and OBS use their own selections.}
\label{tab:coverage_family_fixed}
\end{table}

\clearpage
\begin{table}[H]
\centering\small
\setlength{\tabcolsep}{4pt}
\begin{tabular}{llrlrrrrr}
\toprule
Family & Task & $N$ & Readout & 5\% & 10\% & 15\% & 20\% & 25\% \\
\midrule
A & Triage / navigation & 3095 & Choice & 19 (0.6) & 37 (1.2) & 56 (1.8) & 140 (4.5) & 267 (8.6) \\
 &  &  & OBS & 0 (0.0) & 0 (0.0) & 4 (0.1) & 20 (0.6) & 156 (5.0) \\
\midrule
B & Clinical judgment & 2501 & Choice & 390 (15.6) & 782 (31.3) & 1168 (46.7) & 1427 (57.1) & 1618 (64.7) \\
 &  &  & OBS & 425 (17.0) & 838 (33.5) & 1135 (45.4) & 1438 (57.5) & 1622 (64.9) \\
\midrule
C & Safety review & 1812 & Choice & 10 (0.6) & 25 (1.4) & 38 (2.1) & 85 (4.7) & 143 (7.9) \\
 &  &  & OBS & 12 (0.7) & 27 (1.5) & 60 (3.3) & 134 (7.4) & 220 (12.1) \\
\midrule
D & Physiological assessment & 1334 & Choice & 18 (1.3) & 30 (2.2) & 49 (3.7) & 96 (7.2) & 157 (11.8) \\
 &  &  & OBS & 0 (0.0) & 9 (0.7) & 112 (8.4) & 156 (11.7) & 187 (14.0) \\
\bottomrule
\end{tabular}
\caption{Jev 1.14 coverage by task family under global top-percentage selection. Columns denote global retained proportions; they do not impose equal coverage within tasks. Cells report retained counts followed by within-task coverage (\%) in parentheses. The denominator is the row's $N$, not the total selected pool. Choice and OBS use their own selections.}
\label{tab:coverage_family_fixed_114}
\end{table}

\clearpage
\begin{table}[H]
\centering\small
\setlength{\tabcolsep}{4pt}
\begin{tabular}{llrlrrrrr}
\toprule
Family & Task & $N$ & Readout & $0.70$ & $0.75$ & $0.80$ & $0.85$ & $0.90$ \\
\midrule
A & CARE & 135 & Choice & 62 (45.9) & 53 (39.3) & 41 (30.4) & 31 (23.0) & 25 (18.5) \\
 &  &  & OBS & 23 (17.0) & 22 (16.3) & 19 (14.1) & 15 (11.1) & 11 (8.1) \\
A & ESI & 1000 & Choice & 578 (57.8) & 528 (52.8) & 459 (45.9) & 402 (40.2) & 323 (32.3) \\
 &  &  & OBS & 283 (28.3) & 209 (20.9) & 135 (13.5) & 54 (5.4) & 3 (0.3) \\
A & MIMIC-IV-ED & 1000 & Choice & 576 (57.6) & 493 (49.3) & 428 (42.8) & 365 (36.5) & 298 (29.8) \\
 &  &  & OBS & 247 (24.7) & 191 (19.1) & 134 (13.4) & 55 (5.5) & 5 (0.5) \\
A & Ramaswamy & 960 & Choice & 663 (69.1) & 533 (55.5) & 443 (46.1) & 348 (36.2) & 235 (24.5) \\
 &  &  & OBS & 116 (12.1) & 52 (5.4) & 26 (2.7) & 16 (1.7) & 7 (0.7) \\
\midrule
B & ICD-proxy & 1000 & Choice & 939 (93.9) & 924 (92.4) & 911 (91.1) & 885 (88.5) & 864 (86.4) \\
 &  &  & OBS & 856 (85.6) & 826 (82.6) & 781 (78.1) & 724 (72.4) & 592 (59.2) \\
B & MedCalc & 41 & Choice & 33 (80.5) & 33 (80.5) & 32 (78.0) & 29 (70.7) & 25 (61.0) \\
 &  &  & OBS & 31 (75.6) & 27 (65.9) & 25 (61.0) & 20 (48.8) & 19 (46.3) \\
B & MedQA & 252 & Choice & 215 (85.3) & 201 (79.8) & 191 (75.8) & 183 (72.6) & 169 (67.1) \\
 &  &  & OBS & 170 (67.5) & 146 (57.9) & 126 (50.0) & 99 (39.3) & 54 (21.4) \\
B & MedXpertQA & 130 & Choice & 57 (43.8) & 47 (36.2) & 35 (26.9) & 26 (20.0) & 22 (16.9) \\
 &  &  & OBS & 7 (5.4) & 2 (1.5) & 0 (0.0) & 0 (0.0) & 0 (0.0) \\
B & MIMIC-CDM & 1000 & Choice & 965 (96.5) & 956 (95.6) & 940 (94.0) & 930 (93.0) & 909 (90.9) \\
 &  &  & OBS & 925 (92.5) & 897 (89.7) & 876 (87.6) & 814 (81.4) & 694 (69.4) \\
B & SCT & 78 & Choice & 26 (33.3) & 22 (28.2) & 18 (23.1) & 12 (15.4) & 8 (10.3) \\
 &  &  & OBS & 5 (6.4) & 4 (5.1) & 3 (3.8) & 3 (3.8) & 2 (2.6) \\
\midrule
C & MEDEC & 509 & Choice & 362 (71.1) & 314 (61.7) & 268 (52.7) & 212 (41.7) & 155 (30.5) \\
 &  &  & OBS & 252 (49.5) & 190 (37.3) & 130 (25.5) & 86 (16.9) & 48 (9.4) \\
C & MedVAL & 388 & Choice & 149 (38.4) & 127 (32.7) & 111 (28.6) & 89 (22.9) & 79 (20.4) \\
 &  &  & OBS & 69 (17.8) & 56 (14.4) & 51 (13.1) & 37 (9.5) & 20 (5.2) \\
C & RxSafe & 915 & Choice & 628 (68.6) & 565 (61.7) & 494 (54.0) & 412 (45.0) & 323 (35.3) \\
 &  &  & OBS & 421 (46.0) & 334 (36.5) & 241 (26.3) & 151 (16.5) & 64 (7.0) \\
\midrule
D & CTG-21 & 1000 & Choice & 92 (9.2) & 34 (3.4) & 16 (1.6) & 4 (0.4) & 0 (0.0) \\
 &  &  & OBS & 3 (0.3) & 0 (0.0) & 0 (0.0) & 0 (0.0) & 0 (0.0) \\
D & VitalDB & 334 & Choice & 314 (94.0) & 309 (92.5) & 296 (88.6) & 281 (84.1) & 265 (79.3) \\
 &  &  & OBS & 304 (91.0) & 280 (83.8) & 235 (70.4) & 188 (56.3) & 128 (38.3) \\
\bottomrule
\end{tabular}
\caption{Jev 1.13 coverage by dataset at probability thresholds. Cells report retained counts followed by within-task coverage (\%) in parentheses. The denominator is the row's $N$, not the total selected pool. Choice and OBS use their own selections.}
\label{tab:coverage_dataset_threshold}
\end{table}

\clearpage
\begin{table}[H]
\centering\small
\setlength{\tabcolsep}{4pt}
\begin{tabular}{llrlrrrrr}
\toprule
Family & Task & $N$ & Readout & $0.70$ & $0.75$ & $0.80$ & $0.85$ & $0.90$ \\
\midrule
A & CARE & 135 & Choice & 53 (39.3) & 42 (31.1) & 31 (23.0) & 26 (19.3) & 21 (15.6) \\
 &  &  & OBS & 23 (17.0) & 22 (16.3) & 21 (15.6) & 16 (11.9) & 12 (8.9) \\
A & ESI & 1000 & Choice & 576 (57.6) & 522 (52.2) & 451 (45.1) & 386 (38.6) & 313 (31.3) \\
 &  &  & OBS & 359 (35.9) & 291 (29.1) & 227 (22.7) & 149 (14.9) & 68 (6.8) \\
A & MIMIC-IV-ED & 1000 & Choice & 555 (55.5) & 483 (48.3) & 410 (41.0) & 351 (35.1) & 278 (27.8) \\
 &  &  & OBS & 324 (32.4) & 253 (25.3) & 203 (20.3) & 144 (14.4) & 65 (6.5) \\
A & Ramaswamy & 960 & Choice & 669 (69.7) & 587 (61.1) & 490 (51.0) & 330 (34.4) & 203 (21.1) \\
 &  &  & OBS & 166 (17.3) & 107 (11.1) & 41 (4.3) & 16 (1.7) & 6 (0.6) \\
\midrule
B & ICD-proxy & 1000 & Choice & 937 (93.7) & 919 (91.9) & 899 (89.9) & 879 (87.9) & 848 (84.8) \\
 &  &  & OBS & 885 (88.5) & 856 (85.6) & 815 (81.5) & 773 (77.3) & 716 (71.6) \\
B & MedCalc & 41 & Choice & 35 (85.4) & 32 (78.0) & 28 (68.3) & 28 (68.3) & 24 (58.5) \\
 &  &  & OBS & 31 (75.6) & 30 (73.2) & 26 (63.4) & 23 (56.1) & 22 (53.7) \\
B & MedQA & 252 & Choice & 219 (86.9) & 210 (83.3) & 199 (79.0) & 182 (72.2) & 165 (65.5) \\
 &  &  & OBS & 175 (69.4) & 154 (61.1) & 136 (54.0) & 109 (43.3) & 67 (26.6) \\
B & MedXpertQA & 130 & Choice & 54 (41.5) & 44 (33.8) & 34 (26.2) & 29 (22.3) & 23 (17.7) \\
 &  &  & OBS & 10 (7.7) & 5 (3.8) & 1 (0.8) & 0 (0.0) & 0 (0.0) \\
B & MIMIC-CDM & 1000 & Choice & 968 (96.8) & 952 (95.2) & 941 (94.1) & 923 (92.3) & 900 (90.0) \\
 &  &  & OBS & 943 (94.3) & 926 (92.6) & 905 (90.5) & 865 (86.5) & 808 (80.8) \\
B & SCT & 78 & Choice & 20 (25.6) & 17 (21.8) & 12 (15.4) & 11 (14.1) & 9 (11.5) \\
 &  &  & OBS & 5 (6.4) & 5 (6.4) & 3 (3.8) & 3 (3.8) & 0 (0.0) \\
\midrule
C & MEDEC & 509 & Choice & 320 (62.9) & 269 (52.8) & 224 (44.0) & 171 (33.6) & 123 (24.2) \\
 &  &  & OBS & 270 (53.0) & 216 (42.4) & 172 (33.8) & 108 (21.2) & 56 (11.0) \\
C & MedVAL & 388 & Choice & 135 (34.8) & 105 (27.1) & 79 (20.4) & 64 (16.5) & 58 (14.9) \\
 &  &  & OBS & 59 (15.2) & 50 (12.9) & 44 (11.3) & 34 (8.8) & 14 (3.6) \\
C & RxSafe & 915 & Choice & 624 (68.2) & 545 (59.6) & 467 (51.0) & 394 (43.1) & 291 (31.8) \\
 &  &  & OBS & 483 (52.8) & 395 (43.2) & 305 (33.3) & 228 (24.9) & 143 (15.6) \\
\midrule
D & CTG-21 & 1000 & Choice & 10 (1.0) & 2 (0.2) & 0 (0.0) & 0 (0.0) & 0 (0.0) \\
 &  &  & OBS & 0 (0.0) & 0 (0.0) & 0 (0.0) & 0 (0.0) & 0 (0.0) \\
D & VitalDB & 334 & Choice & 308 (92.2) & 290 (86.8) & 267 (79.9) & 250 (74.9) & 228 (68.3) \\
 &  &  & OBS & 297 (88.9) & 266 (79.6) & 245 (73.4) & 214 (64.1) & 186 (55.7) \\
\bottomrule
\end{tabular}
\caption{Jev 1.14 coverage by dataset at probability thresholds. Cells report retained counts followed by within-task coverage (\%) in parentheses. The denominator is the row's $N$, not the total selected pool. Choice and OBS use their own selections.}
\label{tab:coverage_dataset_threshold_114}
\end{table}

\clearpage
\begin{table}[H]
\centering\small
\setlength{\tabcolsep}{4pt}
\begin{tabular}{llrlrrrrr}
\toprule
Family & Task & $N$ & Readout & 5\% & 10\% & 15\% & 20\% & 25\% \\
\midrule
A & CARE & 135 & Choice & 3 (2.2) & 6 (4.4) & 9 (6.7) & 15 (11.1) & 17 (12.6) \\
 &  &  & OBS & 0 (0.0) & 3 (2.2) & 8 (5.9) & 12 (8.9) & 15 (11.1) \\
A & ESI & 1000 & Choice & 16 (1.6) & 33 (3.3) & 47 (4.7) & 54 (5.4) & 109 (10.9) \\
 &  &  & OBS & 0 (0.0) & 0 (0.0) & 2 (0.2) & 7 (0.7) & 45 (4.5) \\
A & MIMIC-IV-ED & 1000 & Choice & 13 (1.3) & 27 (2.7) & 39 (3.9) & 46 (4.6) & 106 (10.6) \\
 &  &  & OBS & 0 (0.0) & 0 (0.0) & 2 (0.2) & 7 (0.7) & 41 (4.1) \\
A & Ramaswamy & 960 & Choice & 2 (0.2) & 3 (0.3) & 5 (0.5) & 8 (0.8) & 24 (2.5) \\
 &  &  & OBS & 0 (0.0) & 0 (0.0) & 0 (0.0) & 11 (1.1) & 16 (1.7) \\
\midrule
B & ICD-proxy & 1000 & Choice & 154 (15.4) & 301 (30.1) & 454 (45.4) & 609 (60.9) & 710 (71.0) \\
 &  &  & OBS & 184 (18.4) & 373 (37.3) & 507 (50.7) & 621 (62.1) & 719 (71.9) \\
B & MedCalc & 41 & Choice & 4 (9.8) & 7 (17.1) & 8 (19.5) & 10 (24.4) & 16 (39.0) \\
 &  &  & OBS & 9 (22.0) & 13 (31.7) & 16 (39.0) & 19 (46.3) & 19 (46.3) \\
B & MedQA & 252 & Choice & 15 (6.0) & 34 (13.5) & 51 (20.2) & 70 (27.8) & 92 (36.5) \\
 &  &  & OBS & 2 (0.8) & 12 (4.8) & 30 (11.9) & 57 (22.6) & 89 (35.3) \\
B & MedXpertQA & 130 & Choice & 0 (0.0) & 1 (0.8) & 2 (1.5) & 4 (3.1) & 6 (4.6) \\
 &  &  & OBS & 0 (0.0) & 0 (0.0) & 0 (0.0) & 0 (0.0) & 0 (0.0) \\
B & MIMIC-CDM & 1000 & Choice & 181 (18.1) & 353 (35.3) & 534 (53.4) & 715 (71.5) & 787 (78.7) \\
 &  &  & OBS & 223 (22.3) & 425 (42.5) & 606 (60.6) & 718 (71.8) & 805 (80.5) \\
B & SCT & 78 & Choice & 1 (1.3) & 3 (3.8) & 3 (3.8) & 3 (3.8) & 3 (3.8) \\
 &  &  & OBS & 0 (0.0) & 0 (0.0) & 0 (0.0) & 2 (2.6) & 3 (3.8) \\
\midrule
C & MEDEC & 509 & Choice & 3 (0.6) & 10 (2.0) & 19 (3.7) & 25 (4.9) & 39 (7.7) \\
 &  &  & OBS & 16 (3.1) & 24 (4.7) & 33 (6.5) & 52 (10.2) & 79 (15.5) \\
C & MedVAL & 388 & Choice & 8 (2.1) & 18 (4.6) & 24 (6.2) & 26 (6.7) & 37 (9.5) \\
 &  &  & OBS & 0 (0.0) & 4 (1.0) & 15 (3.9) & 25 (6.4) & 35 (9.0) \\
C & RxSafe & 915 & Choice & 11 (1.2) & 19 (2.1) & 27 (3.0) & 33 (3.6) & 72 (7.9) \\
 &  &  & OBS & 3 (0.3) & 19 (2.1) & 37 (4.0) & 72 (7.9) & 139 (15.2) \\
\midrule
D & CTG-21 & 1000 & Choice & 0 (0.0) & 0 (0.0) & 0 (0.0) & 0 (0.0) & 0 (0.0) \\
 &  &  & OBS & 0 (0.0) & 0 (0.0) & 0 (0.0) & 0 (0.0) & 0 (0.0) \\
D & VitalDB & 334 & Choice & 26 (7.8) & 59 (17.7) & 89 (26.6) & 130 (38.9) & 167 (50.0) \\
 &  &  & OBS & 0 (0.0) & 1 (0.3) & 55 (16.5) & 145 (43.4) & 180 (53.9) \\
\bottomrule
\end{tabular}
\caption{Jev 1.13 coverage by dataset under global top-percentage selection. Across datasets, each column retains 437, 874, 1,311, 1,748, or 2,185 entries per readout. Cells report retained counts followed by within-task coverage (\%) in parentheses. The denominator is the row's $N$, not the total selected pool. Choice and OBS use their own selections.}
\label{tab:coverage_dataset_fixed}
\end{table}

\clearpage
\begin{table}[H]
\centering\small
\setlength{\tabcolsep}{4pt}
\begin{tabular}{llrlrrrrr}
\toprule
Family & Task & $N$ & Readout & 5\% & 10\% & 15\% & 20\% & 25\% \\
\midrule
A & CARE & 135 & Choice & 1 (0.7) & 2 (1.5) & 7 (5.2) & 10 (7.4) & 12 (8.9) \\
 &  &  & OBS & 0 (0.0) & 0 (0.0) & 0 (0.0) & 6 (4.4) & 12 (8.9) \\
A & ESI & 1000 & Choice & 9 (0.9) & 17 (1.7) & 24 (2.4) & 60 (6.0) & 107 (10.7) \\
 &  &  & OBS & 0 (0.0) & 0 (0.0) & 1 (0.1) & 7 (0.7) & 70 (7.0) \\
A & MIMIC-IV-ED & 1000 & Choice & 9 (0.9) & 17 (1.7) & 24 (2.4) & 61 (6.1) & 117 (11.7) \\
 &  &  & OBS & 0 (0.0) & 0 (0.0) & 3 (0.3) & 7 (0.7) & 68 (6.8) \\
A & Ramaswamy & 960 & Choice & 0 (0.0) & 1 (0.1) & 1 (0.1) & 9 (0.9) & 31 (3.2) \\
 &  &  & OBS & 0 (0.0) & 0 (0.0) & 0 (0.0) & 0 (0.0) & 6 (0.6) \\
\midrule
B & ICD-proxy & 1000 & Choice & 168 (16.8) & 328 (32.8) & 501 (50.1) & 622 (62.2) & 712 (71.2) \\
 &  &  & OBS & 91 (9.1) & 340 (34.0) & 479 (47.9) & 642 (64.2) & 718 (71.8) \\
B & MedCalc & 41 & Choice & 6 (14.6) & 10 (24.4) & 12 (29.3) & 15 (36.6) & 19 (46.3) \\
 &  &  & OBS & 8 (19.5) & 10 (24.4) & 15 (36.6) & 17 (41.5) & 22 (53.7) \\
B & MedQA & 252 & Choice & 22 (8.7) & 43 (17.1) & 60 (23.8) & 78 (31.0) & 103 (40.9) \\
 &  &  & OBS & 1 (0.4) & 2 (0.8) & 7 (2.8) & 27 (10.7) & 69 (27.4) \\
B & MedXpertQA & 130 & Choice & 0 (0.0) & 1 (0.8) & 4 (3.1) & 6 (4.6) & 8 (6.2) \\
 &  &  & OBS & 0 (0.0) & 0 (0.0) & 0 (0.0) & 0 (0.0) & 0 (0.0) \\
B & MIMIC-CDM & 1000 & Choice & 193 (19.3) & 397 (39.7) & 588 (58.8) & 703 (70.3) & 772 (77.2) \\
 &  &  & OBS & 325 (32.5) & 486 (48.6) & 634 (63.4) & 752 (75.2) & 812 (81.2) \\
B & SCT & 78 & Choice & 1 (1.3) & 3 (3.8) & 3 (3.8) & 3 (3.8) & 4 (5.1) \\
 &  &  & OBS & 0 (0.0) & 0 (0.0) & 0 (0.0) & 0 (0.0) & 1 (1.3) \\
\midrule
C & MEDEC & 509 & Choice & 2 (0.4) & 8 (1.6) & 12 (2.4) & 27 (5.3) & 37 (7.3) \\
 &  &  & OBS & 9 (1.8) & 12 (2.4) & 23 (4.5) & 38 (7.5) & 57 (11.2) \\
C & MedVAL & 388 & Choice & 2 (0.5) & 5 (1.3) & 6 (1.5) & 19 (4.9) & 29 (7.5) \\
 &  &  & OBS & 0 (0.0) & 0 (0.0) & 0 (0.0) & 4 (1.0) & 17 (4.4) \\
C & RxSafe & 915 & Choice & 6 (0.7) & 12 (1.3) & 20 (2.2) & 39 (4.3) & 77 (8.4) \\
 &  &  & OBS & 3 (0.3) & 15 (1.6) & 37 (4.0) & 92 (10.1) & 146 (16.0) \\
\midrule
D & CTG-21 & 1000 & Choice & 0 (0.0) & 0 (0.0) & 0 (0.0) & 0 (0.0) & 0 (0.0) \\
 &  &  & OBS & 0 (0.0) & 0 (0.0) & 0 (0.0) & 0 (0.0) & 0 (0.0) \\
D & VitalDB & 334 & Choice & 18 (5.4) & 30 (9.0) & 49 (14.7) & 96 (28.7) & 157 (47.0) \\
 &  &  & OBS & 0 (0.0) & 9 (2.7) & 112 (33.5) & 156 (46.7) & 187 (56.0) \\
\bottomrule
\end{tabular}
\caption{Jev 1.14 coverage by dataset under global top-percentage selection. Across datasets, each column retains 437, 874, 1,311, 1,748, or 2,185 entries per readout. Cells report retained counts followed by within-task coverage (\%) in parentheses. The denominator is the row's $N$, not the total selected pool. Choice and OBS use their own selections.}
\label{tab:coverage_dataset_fixed_114}
\end{table}
\begin{multicols}{2}
\section{Ethical Considerations}

This study evaluates the reliability of automated decision models in medical settings, where incorrect or overconfident predictions may have serious consequences for patient safety. Our experiments are conducted on existing medical benchmarks and are intended for research evaluation rather than clinical deployment. The evaluated models should not be used as substitutes for qualified healthcare professionals, particularly given the observed limitations in numerical assessment, abstention awareness, and confidence-based decision selection. Although our analysis identifies potential improvements in reliability, stronger benchmark performance does not establish clinical safety or readiness for real-world use. Furthermore, medical datasets may contain demographic imbalances, annotation uncertainties, and biases inherited from their original sources, which could affect model behavior across patient populations. The use of clinical data also requires appropriate attention to privacy, data access conditions, and responsible handling of potentially sensitive information. We therefore emphasize that further validation, including assessments of population-level disparities and prospective clinical evaluation, is necessary before considering these models for consequential medical applications.
\end{multicols}
\end{document}